\documentclass[lettersize,journal]{IEEEtran}
\usepackage{amsmath,amsfonts}
\usepackage{algorithmic}
\usepackage{algorithm}
\usepackage{array}
\usepackage[caption=false,font=normalsize,labelfont=sf,textfont=sf]{subfig}
\usepackage{textcomp}
\usepackage{stfloats}
\usepackage{url}
\usepackage{verbatim}
\usepackage{graphicx}
\usepackage{cite}
\usepackage{subcaption}
\begin{document}

\title{Novel Actor-Critic Algorithm for Robust Decision Making of CAV under Delays and Loss of V2X Data}

\author{Zine~el~abidine~Kherroubi
\thanks{Z. Kherroubi is with the Technology Innovation Institute, 9639 Masdar City, Abu Dhabi, United Arab Emirates (emails: zine.kherroubi@tii.ae).}
}

\markboth{SUBMITTED TO IEEE TRANSACTIONS ON INTELLIGENT TRANSPORTATION SYSTEMS}%
{Shell \MakeLowercase{\textit{et al.}}: A Sample Article Using IEEEtran.cls for IEEE Journals}

\IEEEpubid{}

\maketitle

\begin{abstract}
Current autonomous driving systems heavily rely on V2X communication data to enhance situational awareness and the cooperation between vehicles. However, a major challenge when using V2X data is that it may not be available periodically because of unpredictable delays and data loss during wireless transmission between road stations and the receiver vehicle. This issue should be considered when designing control strategies for autonomous and connected vehicles. This paper proposes a novel \textsl{‘Blind Actor-critic’} algorithm that guarantees robust driving performance in V2X environment with delayed and/or lost data. The novel algorithm incorporates three keys mechanisms: a virtual fixed sampling period, a combination of \textsl{Temporal-Difference} and \textsl{Monte Carlo} learning, and the numerical approximation of immediate reward values. To address the \textsl{temporal aperiodicity problem} of V2X data, we first illustrate this challenge. Then, we provide a detailed explanation of the \textsl{Blind Actor-critic} algorithm where we highlight the proposed components to compensate for the temporal aperiodicity problem of V2X data. We evaluate the performance of our algorithm in a simulation environment and compare it to benchmark approaches. The results demonstrate that training metrics are improved compared to conventional Actor-critic algorithms. Additionally, testing results show that our approach provides robust control, even under low V2X network reliability levels.
\end{abstract}

\section{Introduction}\label{intro}
\IEEEPARstart{R}{search} interest in Connected and Autonomous Vehicle (CAV) technologies is increasing because of their abilities to improve driving safety and efficiency. While classic driving assistance systems rely only on onboard vehicle's sensors that are limited in range and Field-of-View (FoV), sharing vehicles' data allows extending perception range, overcoming blind spots and Line-Of-Sight (LOS) obstacles, and increasing detection accuracy. To achieve this, Vehicle-To-everything (V2X) communication is the enabling technology. 
Through V2X communication, vehicles can exchange data directly with other vehicles (V2V), with road-side infrastructure (V2I), and even with pedestrians (V2P). According to the U.S. Department of Transportation (USDOT), about 82\% of accidents involving unimpaired drivers can be addressed by the successful deployment of connected vehicle (CV) technology \cite{CV_crash_rate_1,CV_crash_rate_2}. Connected and autonomous vehicles could meet their strict functional requirements using V2X applications of cooperative awareness and collective perception. Cooperative awareness enables vehicles to transmit data about their state (current position, speed, and heading, etc.) via V2X communication \cite{ETSI_CAM,SAE_CAM}, while collective perception allows Intelligent Transport Systems Stations (ITS-Ss), including vehicles and road infrastructure, to inform other connected ITS-Ss of objects (such as pedestrians, obstacles, and other vehicles) detected by their local sensors \cite{ETSI_CPM}. It was already shown that collective perception guarantees higher object awareness ratio compared to onboard perception \cite{ETSI_CPM}, and improves the object detection accuracy \cite{CPM_accuracy_1,CPM_accuracy_2,CPM_accuracy_3}.
\subsection{Objectives and contributions}\label{objmotiv}
Regarding its capabilities, V2X technology has indeed become a key enabler in current autonomous driving systems. However, delays and loss in V2X networks are still an up-to-date challenges when using this technology for safety-critical applications, which are very real-time sensitives. To address this, we introduce a novel algorithm, \textsl{Blind Actor-critic}, designed to mitigate the \textsl{temporal aperiodicity problem}, caused by delays and loss, when using V2X data for controlling CAV. Indeed, deep reinforcement learning (DRL) has demonstrated its performance in many continuous control applications \cite{DRL}, including autonomous driving \cite{Ping_Wang,Kherroubi2022}. Furthermore, the Actor-Critic family of algorithms combines both value-based and policy-based methods to offer a balanced approach for solving various continuous control problems \cite{Sutton}. Nonetheless, most of these DRL techniques assume full availability of environment information (state and reward in the reinforcement learning terminology) at constant sampling frequency. This contrasts with delays and data loss that are presents in telecommunication networks in general, and V2X networks in particular. To the best of our knowledge, the approach presented in \cite{Delayed_AC} is the most recent work that addresses random delays based on an Actor-critic algorithm. While it was successfully tested and validated in a \textsl{MuJoCo}\cite{Mujoco} continuous control benchmark environment, this approach makes the hard assumption of \textsl{knowing the dynamics of the environment}. This condition could not be satisfied in practice, which makes the applicability of that approach highly questionable.\\
\indent In this paper, we introduce a novel approach for controlling autonomous driving using V2X data as a primary source of information. Our contributions include:
\begin{itemize}
\item A novel Actor-critic-based algorithm, \textsl{Blind Actor-critic}, specifically designed to overcome temporal aperiodicity problem when using V2X data for vehicle control.
\item A new improvements on the classic Actor-critic paradigm that could be used when the environment's information (state and reward) is subject to delays or loss within the communication network.
\item Implementation, testing, and validation of this new approach in a highway-on ramp merging use case within a simulated driving environment.
\item Demonstrated improvements in both the training and testing performance compared to the classic Actor-critic algorithm. \item The novel approach ensures robust vehicle driving control when utilizing V2X data, even when these data are delayed and lost within the V2X network.
\end{itemize}
\subsection{Paper Organization}\label{organization}
The paper is organized as follows: In next Section \ref{chlgmtv}, we introduce and illustrate the problem of temporal aperiodicity of V2X data that will be addressed by our approach. Then, Section \ref{SOTA} examines related works dealing with temporal aperiodicity in V2X communication. Section \ref{AC} provides an overview of the Actor-critic family of deep reinforcement learning techniques. In Section \ref{methodology}, we deep dive into a detailed description of our proposed approach. Subsequently, in Section \ref{Sim_framework}, we introduce the simulation framework which is utilized in our study, along with the highway on-ramp merging use case used for test and validation. Section \ref{Res_disc} is dedicated to presenting and discussing the obtained results. Finally, in Section \ref{conc}, we draw our concluding findings and further perspectives. 
\section{Challenges and motivations}\label{chlgmtv}
\begin{figure}[b]
\centering
\includegraphics[scale=0.36]{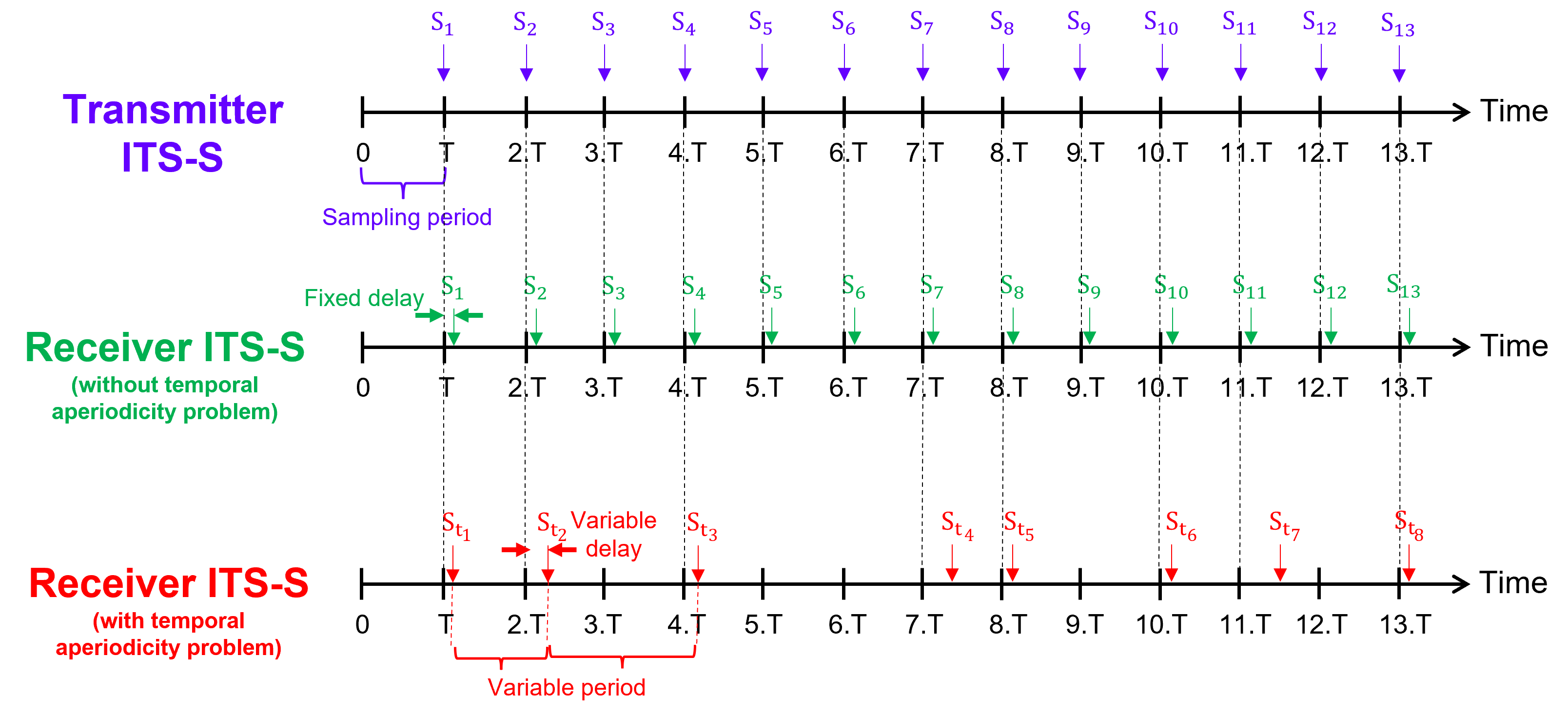}
\caption{Illustration of \textsl{the temporal aperiodicity} problem when using V2X data: environment state $(S_i~:~i \in N)$ is perceived by the transmitter ITS-S with a fixed sampling period (T). In the ideal case, the receiver ITS-S receives the environment state with the same fixed sampling period (T), as shown by green color. In practice, the receiver ITS-S receives the environment state asynchronously at varied instants $(S_{t_i}~:~t_i \in R^+)$, as shown in red.}
\label{fig_temp_aperiod}
\end{figure}

\begin{figure}[!b]
\centering
\subfloat[Use case with V2V communication]{\includegraphics[scale=0.22]{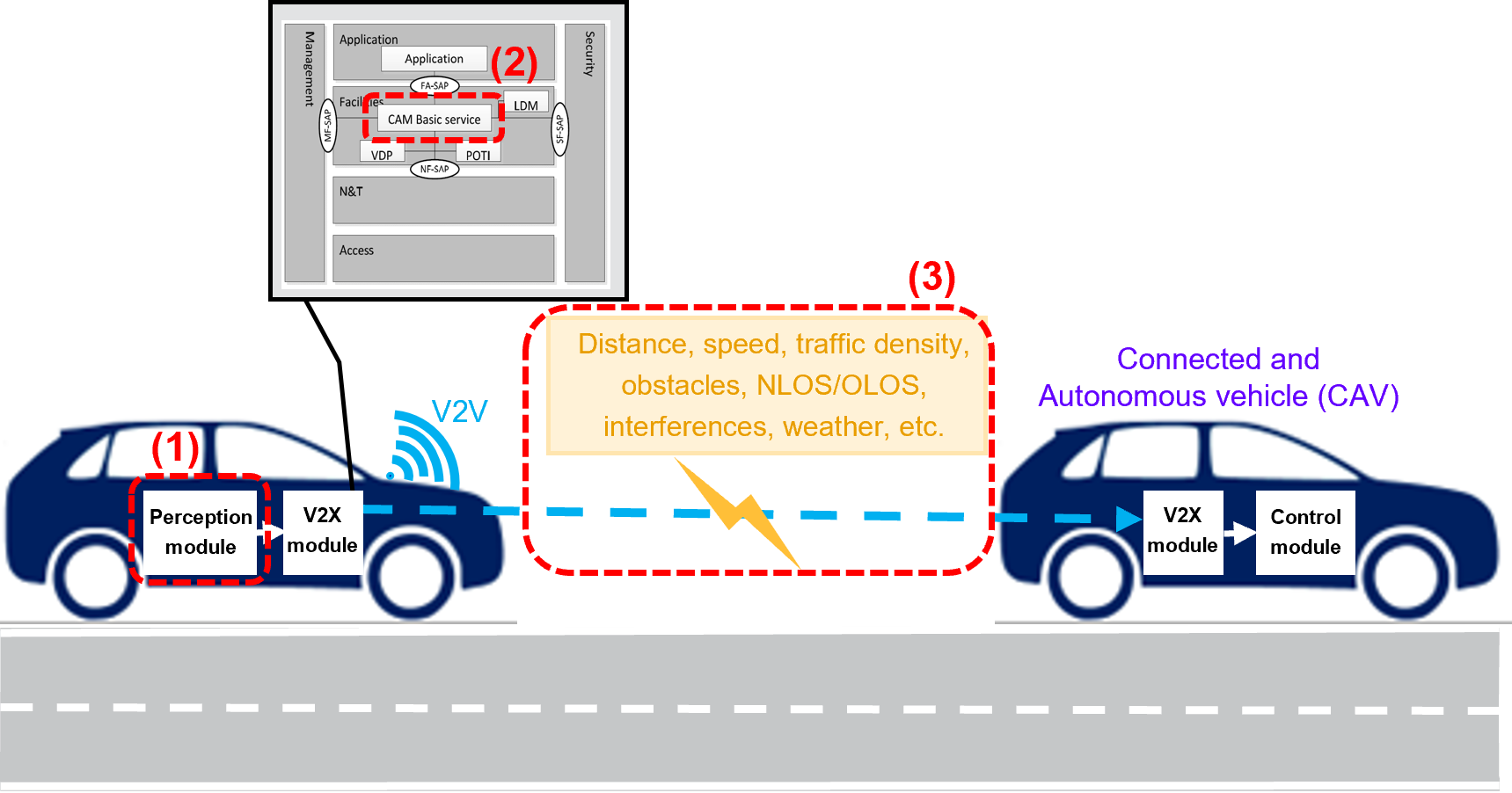}}%
\vfil
\subfloat[Use case with V2I communication]{\includegraphics[scale=0.22]{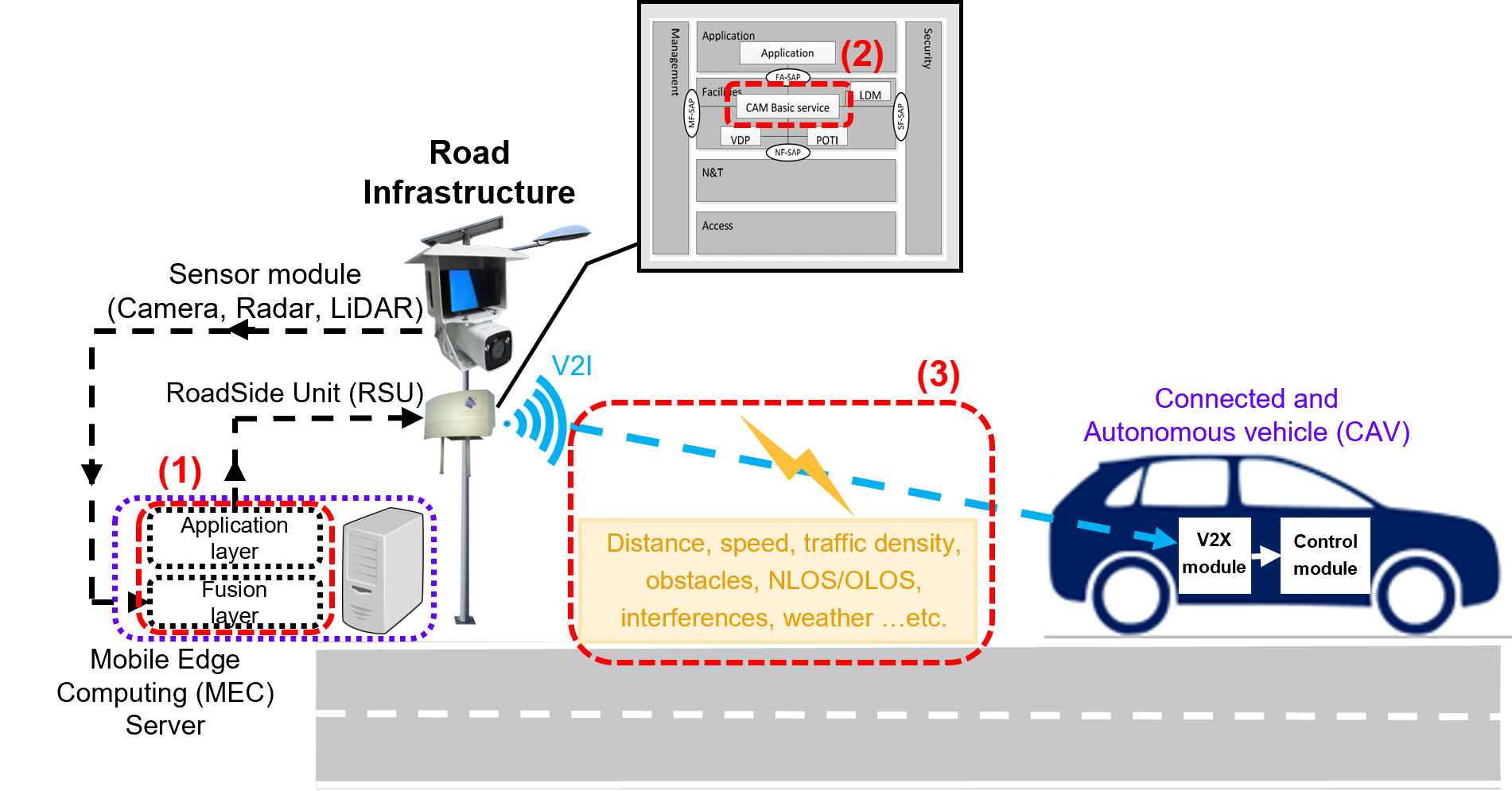}}%
\caption{Illustration of two use cases where temporal aperiodicity factors are mentioned by dashed red boxes \textsl{(a)} use case with V2V communication and \textsl{(b)} use case with V2I communication.}
\label{fig_vv_vs_vi}
\end{figure}

Due to its ability to enhance situational awareness, V2X technology has emerged as an essential enabler for autonomous driving applications. Significant research and development efforts are being invested to standardize and deploy this technology at scale. To cite, cooperative awareness has already been standardized, which basically relies on Cooperative Awareness Message (CAM) in Europe \cite{ETSI_CAM} and the Basic Safety Message (BSM) in the U.S. \cite{SAE_CAM}. Furthermore, collective perception is currently in its final stages of standardization in both Europe \cite{ETSI_CPM} and the U.S. \cite{SAE_CPM}.\\
To fully leverage the full potential of V2X, a wide range of applications and requirements have been defined \cite{ETSI_V2X_app}. The primary applications of V2X are road safety use cases. These use cases include control and decision applications which are very real time critical. However, despite the increased informational and situational awareness offered by V2X, a significant challenge arises when using V2X information for decision and control of CAV. Indeed, V2X communication messages are provided by external ITS-Ss (other connected vehicles and roadside units) that are not controlled or synchronized directly by the ego CAV. Consequently, V2X data may be received at unpredictable instants of time and in an asynchronous manner. We refer to this challenge as \textsl{temporal aperiodicity problem}, as depicted in Fig. \ref{fig_temp_aperiod}. This problem becomes particularly relevant when using V2X data as input to autonomous driving systems, in contrast to the sole reliance on onboard sensors, which operate in sync with the vehicle's internal system clock and are directly controlled and accessed by the ego CAV.
In practice, three main factors contribute to temporal aperiodicity within a V2X network:
\begin{itemize}
\item \textbf{Processing delay:} This is the time required to process perception data by the transmitter Intelligent Transportation System Station (ITS-S). The processing time is unknown to the receiver ITS-S and could vary depending on factors such as the type of sensors used (camera, radar, LiDAR, etc.), the algorithms executed (data preprocessing, fusion algorithm, etc.), and the processing hardware utilized. 
\item \textbf{V2X messages generation rules:} The generation rules for V2X messages, as defined by standards, are dynamic and follow specific criteria \cite{ETSI_CAM}\cite{ETSI_CPM}. To illustrate, Collective Perception Messages (CPM) are generated and transmitted under the following conditions \cite{ETSI_CPM}:
\begin{itemize}
\item A new object has been detected since the last message transmission.
\item The time elapsed since the last message transmission falls within the range of 0.1 seconds to 1 second.
\item The difference in the object's position since the last message transmission exceeds 4 meters.
\item The difference in the object's speed since the last message transmission exceeds 0.5 meters per second.
\item The difference in the object's heading since the last message transmission exceeds 4 degrees.
\end{itemize}
This implies that V2X messages are not generated synchronously at fixed frequency. These rules for generating V2X messages represent a trade-off between update frequency and channel load to preserve the stability and performance of the V2X network.
\item \textbf{V2X network reliability:} The major causes of temporal aperiodicity problem are latency and data loss in the V2X network, depending on its reliability. Communication reliability level is subject to various dynamic factors, including technology type (DSRC vs C-V2X) \cite{V2X_delays_reference}\cite{DSRC_LTE}, distance between the ITS-Ss, speed of vehicles, traffic density, environment geometry, presence of obstacles, Line-Of-Sight (LOS) conditions, the presence of interference sources, weather conditions, and more \cite{Wireless_book}.
\end{itemize}
These factors are illustrated in Fig. \ref{fig_vv_vs_vi} for use cases with V2V and V2I communication.\\
\indent Temporal aperiodicity problem was previously ignored in the literature when using V2X communication for safety related applications, like decision and control. However, controlling autonomous vehicle requires real-time update of both the vehicle's and environment's state. Unlike conventional driving systems where the information is guaranteed reliably by on-board sensors, systems relying on V2X data for situational awareness may encounter delays and data loss due to the aforementioned factors (as depicted in Fig. \ref{fig_vv_vs_vi}). Consequently, the performance of such systems may drop or even deteriorate. Indeed, it has been demonstrated that V2I communication distortions can impact the optimal control of CAVs, potentially resulting in lateral collisions \cite{V2X_delays_reference}. Even breakthrough AI techniques, such as deep reinforcement learning, may see their performance drop when utilizing V2X data, as will be demonstrated later in this paper.
\section{Related works}\label{SOTA}
Most works that studied V2X safety-related applications assume a highly reliable (i.e. perfect) communication link. However, this assumption is not realistic, as was illustrated before. Communication distortions and imperfect data transmission have been previously ignored in the literature, and only few studies considered the bi-directional dependency between V2X network reliability level and driving performance.\\
Earlier work \cite{CCA_V2X} showed that highway cooperative collision avoidance (CCA) system may cause longitudinal collision between adjacent vehicles below some wireless channels reliability level (i.e. some latency and packet loss thresholds). A subsequent study \cite{HU2019506} uncovered that communication delays deteriorate both dynamic stability and static performance of a platoon of adaptive cruise control (ACC) vehicles. Comparatively, the negative effect from communication delay is larger than that from actuator lag. Beside that, another work \cite{WANG2018276} proved that deviation in the feasible control domain (the domain where control variable(s) satisfies (satisfy) all pre-defined constrains), caused by communication delays, may alter or narrow the stability domain of the control system. Therefore, in order to address this issue, a dynamic controller manager proposed in \cite{platoon_CACC} monitors the channel quality within the platoon of (ACC) vehicles and reacts by degrading performance to preserve safety in the presence of communication failures. However, the study in Ref. \cite{V2X_delays_reference} showed that even optimal control of vehicles may lead to potential lateral collision when V2I data is used with communication delays. To overcome this challenge, authors of \cite{motion_estimation} proposed a consensus-based motion estimation methodology to estimate the vehicle motion when the vehicular communication network is not reliable. This approach was tested in a simulation environment for an unsignalized intersection at medium speed of 50 km/h, where the position estimation error was within the range $\pm 0.5 m$. Nevertheless, estimation methods may incur approximation errors due to non-linearity of vehicle's dynamic (inertia, frictions, drag resistance, etc.), especially at high speed. These errors may cause control algorithms to provide unsafe actions or may be accumulated and propagated during learning phase (e.g., through the \textsl{Bellman} equation when using reinforcement learning). The system proposed in Ref. \cite{Patent_1} uses an internal-delay estimation mechanism for each driving scenario to compensate for these delays when controlling autonomous vehicles. Mainly, these delays concern only actuators lag. Such mechanism cannot be applied to compensate for V2X communication delays and loss because accurately modeling them in real-world scenarios, where V2X network performance is subject to variations (e.g., traffic density, vehicle speed, weather conditions), can be a complex and often infeasible task. Another solution, as described in Ref. \cite{patent_2}, involves a vehicle remote control system that is assisted by the roadside unit (i.e. through V2I communication). V2I communication degradation may also cause delayed remote command and could potentially lead to safety issues. The proposed approach in \cite{patent_2} suggests to use redundant devices, such as geographically adjacent roadside units, to guarantee reliable communication (i.e.  low communication delay, high throughput, etc.) when remotely controlling vehicles. However, this solution is not feasible in practice because of economical cost. Beyond safety-critical applications, V2X communication distortions impact also traffic management applications. Indeed, simulation studies \cite{Traffic_control} have demonstrated that adaptive intersection control, a traffic management application, may suffer from significantly increased average delays of vehicles because of communication distortions.\\
\indent In recent years, deep reinforcement learning techniques (i.e. Artificial Intelligence) has proven its performance in many continuous control applications \cite{DRL}, including autonomous driving \cite{Ping_Wang,Kherroubi2022}. However, most of these methods assume observation of environment states and rewards at fixed periods. We demonstrate that performance of classic Actor-critic techniques drops when input information (i.e. states and rewards) is provided through V2X network with delays and data loss. Most reinforcement learning approaches that target \textsl{system delays} consider only delayed reward problems (see subsection \textsl{2.9} in Ref. \cite{RL_challenges}). These techniques deal only with the credit assignment problem \cite{Credit_assign}. On another side, reinforcement learning methods with sparse rewards deal with agents that receive rewards when tasks are completely or partially completed while environment states is fully available. It is obvious that temporal aperiodicity in V2X data (see. Fig. \ref{fig_temp_aperiod}) is more challenging than a credit assignment problem or a sparse rewards setting. Authors of \cite{Blind_RL} proposed a reinforcement learning approach with delayed observations that maximizes the expected returns across all possible future states conditioned over the last known state, and actions taken. Similarly, Ref. \cite{Delayed_AC} proposes a Delay-Correcting Actor-Critic (DCAC), an algorithm based on Soft Actor-Critic for environments with delays. Recently, authors of \cite{Delayed_DQN} proposed a deep Q-network (DQN) algorithm with a forward model to perform in random-delayed 5G networks. The forward model iteratively predicts the next state using ground-truth state and action. These approaches \cite{Blind_RL,Delayed_AC,Delayed_DQN} have a main common limitation: they are a model-based methods (i.e. non model-free RL), which means that they rely on the model of the environment to maximize rewards. In practical applications, real-world environment model is not available and cannot be estimated accurately. For example, dynamic models of surrounding vehicles (mass, inertia, non-linear frictions, etc.) are unknown for an autonomous vehicle in a driving scenario.\\
\indent In the following, we propose a novel model-free Actor-critic algorithm to overcome delays and data loss in communication networks in general, and V2X networks in particular.
\section{Preliminaries on Actor-Critic Methods}\label{AC}
\begin{figure}[t]
\centering
\includegraphics[scale=0.52]{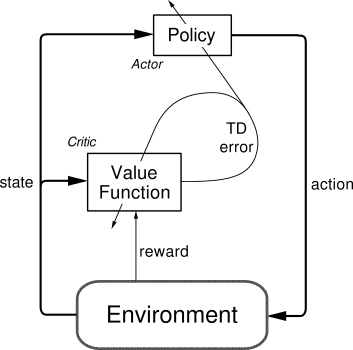}
\caption{The Actor-critic architecture \cite{Sutton}.}
\label{AC_arch}
\end{figure}

As defined in \cite{Sutton}, Actor-critic methods are \textsl{Temporal-Difference} (TD) methods that have a separate memory structure to explicitly represent the policy $\pi_\theta$, parameterized by $\theta$, independent of the value function $V_\phi$, parameterized by $\phi$. The policy structure is known as the actor, because it is used to select actions, and the estimated value function is known as the critic, because it criticizes the actions made by the actor. Learning is always on-policy: the critic must learn about and critique whatever policy is currently being followed by the actor. The critique takes the form of a TD error. This scalar signal is the sole output of the critic and drives all learning in both actor and critic, as suggested by Fig. \ref{AC_arch}.
Actor-critic methods are the natural extension of the idea of reinforcement comparison methods (section 2.8 in \cite{Sutton}) to TD learning and to the full reinforcement learning problem. Typically, the critic is a state-value function. The Actor-critic algorithm is summarized in \textbf{Algorithm 1}. After each action selection (line \textsl{3}), the critic evaluates the new state (line \textsl{5}) to determine whether things have gone better or worse than expected. That evaluation is the TD error (line \textsl{6}):
\begin{equation}
\delta_i = r_{i} + \gamma V_\phi(s_{i+1}) - V_\phi(s_i),
\end{equation}
where $V_\phi$ is the current value function implemented by the critic. This TD error can be used to evaluate the action just selected, the action $a_i$ taken in state $s_i$ (line \textsl{7}). If the TD error is positive, it suggests that the tendency to select  should be strengthened for the future, whereas if the TD error is negative, it suggests the tendency should be weakened (line \textsl{8}).
\begin{table}[ht]
\centering
\begin{tabular}{l}
\hline
\multicolumn{1}{c}{\textbf{Algorithm 1} \textsl{Actor-Critic algorithm}}\\ \hline
\multicolumn{1}{|l|}{\begin{tabular}[c]{@{}l@{}}
{\scriptsize \textsl{1:} \textbf{for} episode = 1, M \textbf{do}}\\
{\scriptsize \textsl{2:} \hspace*{2mm} \textbf{for} i = 1, N \textbf{do}}\\
{\scriptsize \textsl{3:} \hspace*{5mm} Take action $a_{i} \sim \pi_\theta(a|s_{i})$.}\\
{\scriptsize \textsl{4:} \hspace*{5mm} Get next state $s_{i+1}$ and observe reward $r_{i} = R(s_{i+1})$.\;\;\;\;\;\;\;\;\;\;\;\;\;\;\;\;\;\;\;\;\;\;\;}\\
{\scriptsize \textsl{5:} \hspace*{5mm} Update $\hat{V}^\pi_\phi$ using target $\hat{V}^\pi_{tar}$
:}\\
{\scriptsize \hspace*{10mm} $\hat{V}^\pi_{tar} (s_i) = r_i + \gamma \hat{V}^\pi_{tar}(s_{i+1})$}\\
{\scriptsize \textsl{6:} \hspace*{5mm} Evaluate $\hat{A}^\pi (s_i,a_i) = \hat{V}^\pi_{tar}(s_i,a_i) - \hat{V}^\pi_\phi(s_i,a_i)$}\\
{\scriptsize \textsl{7:} \hspace*{5mm} Evaluate $\nabla_\theta J(\theta) \approx \nabla_\theta \log \pi_\theta(a|s) \hat{A}^\pi(s,a)$}\\
{\scriptsize \textsl{8:} \hspace*{5mm} Update $\theta \leftarrow \theta + \alpha \nabla_\theta J(\theta)$}\\
{\scriptsize \textsl{9:} \hspace*{2mm} \textbf{end for}}\\
{\scriptsize \textsl{10:} \textbf{end for}}\\
\end{tabular}}
\\ \hline
\end{tabular}
\end{table}
\section{Methodology}\label{methodology}
\begin{figure}[t]
\centering
\includegraphics[scale=0.4]{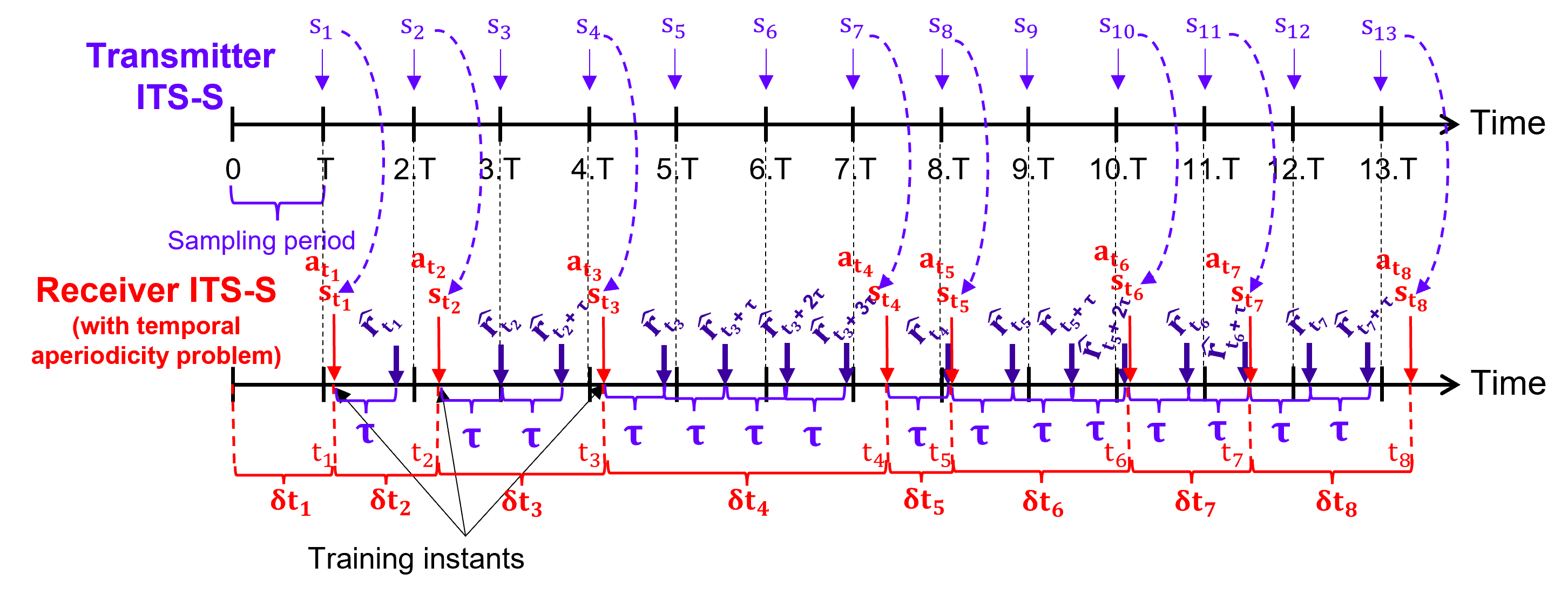}
\caption{Illustration of introduced variables: fictive sampling period \textbf{$\tau$}, random delay $\delta t_i$, and approximate reward $\hat{r}_{t_i + k.\tau}$.}
\label{Temp_schema}
\end{figure}

We propose a novel model-free Actor-critic algorithm, \textsl{Blind Actor-critic}, to use with non-periodic data (i.e. states and rewards). This new approach is built on top of classic Actor-critic algorithm by incorporating three main improvements:
\begin{itemize}
\item First, we introduce a fictive factor \textbf{$\tau$} that forces the Actor-critic algorithm to \textsl{virtually} observe a fixed sampling period of \textbf{$\tau$}, even if input data (i.e. states and rewards) are received with temporal aperiodicity (i.e. with delays and/or loss). This factor could be considered a new hyper-parameter when training the algorithm.
\item Second, we combine \textsl{Temporal-Difference} learning when ground-truth state $s_{t_i}$ is received at random delay $\delta t_i$ with \textsl{Monte Carlo} learning along the path ($t_i$, $t_i + \tau$, $t_i + 2.\tau$, $t_i + 3.\tau$,..., $t_i+\delta t_{i+1}$) (i.e. when ground-truth state is not received).
\item Third, we approximate the immediate reward ($\hat{r}_{t_i}$, $\hat{r}_{t_i + \tau}$, $\hat{r}_{t_i + 2.\tau}$, $\hat{r}_{t_i + 3.\tau}$,..., $\hat{r}_{t_i+\delta t_{i+1}}$) along the path ($t_i$, $t_i + \tau$, $t_i + 2.\tau$, $t_i + 3.\tau$,..., $t_i+\delta t_{i+1}$) using \textsl{n} order approximation, where \textsl{n} depends on the design of the reward function \textsl{R(s)} $\in$ $C^n$.
\end{itemize}
These variables are illustrated in Fig. \ref{Temp_schema}.\\
\indent The primary concept behind these mechanisms is to force the algorithm to mimic an ideal scenario where input data (i.e. states and rewards) are received periodically at a fixed \textbf{$\tau$} period. The algorithm compensates for delayed and lost input data (i.e. states and rewards) using approximate rewards with \textsl{Monte Carlo} learning. Rewards approximation is performed \textsl{numerically} and, hence, no environment model is required. The \textsl{Blind Actor-critic} algorithm is summarized in \textbf{Algorithm 2}. Whenever input data is received through the communication network (i.e. V2X network) at non-periodic instants ($t_i$) (line \textsl{2}), training is performed only if the delay is greater than the fictive sampling period $\tau$ (line \textsl{3}). Actions are executed according to the policy (line \textsl{4}). The ground truth of next state and immediate reward is obtained at instant $t_{i+1}$, after $\delta t_{i+1}$ delay (line \textsl{5}). The value function is updated using \textsl{Temporal-Difference} and \textsl{Monte Carlo} learning (line \textsl{6}). \textsl{Temporal-Difference} is performed using a modulated discount factor ($\gamma^{(\frac{\delta t_{i+1}}{\tau})}$), while \textsl{Monte Carlo} learning is performed using approximate rewards (line \textsl{8}), where these approximations are performed numerically (line \textsl{9}). The advantage is then evaluated (line \textsl{10}) and used to calculate the gradient \textsl{w.r.t} to policy parameters in order to maximize the expected rewards (line \textsl{11}). Finally, policy parameters are updated (line \textsl{12}).
\begin{table}[hb!]
\centering
\begin{tabular}{l}
\hline
\multicolumn{1}{c}{\textbf{Algorithm 2} \textsl{\textsl{Blind Actor-critic} algorithm}}\\ \hline
\multicolumn{1}{|l|}{\begin{tabular}[c]{@{}l@{}}
{\scriptsize \textsl{1:} \textbf{for} episode = 1, M \textbf{do}}\\
{\scriptsize \textsl{2:} \hspace*{1mm} \textbf{for} $t_i = t_1$, $t_N \in \mathbb{R}^*_+$ \textbf{do}}\\
{\scriptsize \textsl{3:} \hspace*{3mm} \textbf{if} $\delta t_i \geq \tau$ \textbf{do}}\\
{\scriptsize \textsl{4:} \hspace*{4mm} Take action $a_{t_{i}} \sim \pi_\theta(a|s_{t_{i}})$.}\\
{\scriptsize \textsl{5:} \hspace*{4mm} Get next state $s_{t_{i+1}}$ and observe reward $r_{t_{i}} = R(s_{t_{i+1}})$.}\\
{\scriptsize \textsl{6:} \hspace*{4mm} Update $\hat{V}^\pi_\phi$ using target $\hat{V}^\pi_{tar}$:}\\
{\scriptsize \hspace*{6mm} $\hat{V}^{\pi}_{\text{tar}}(S_{t_{i}}) = \left[\sum_{k=0}^{\text{int}\left(\frac{\delta t_{i+1}}{\tau}\right) - 1} \gamma^k \cdot \hat{r}_{t_i+k\tau} \right] + \gamma^{\left(\frac{\delta t_{i+1}}{\tau}\right)} \cdot \hat{V}^{\pi}_{\text{tar}}(S_{t_{i+1}})$}\\
{\scriptsize \hspace*{10mm} where:}\\
{\scriptsize \textsl{7:} \hspace*{6mm} $\bullet$ $\delta t_{i+1}$: Time between receiving $s_{t_{i}}$ and $s_{t_{i+1}}$; $\delta t_{i+1} \geq \tau$}\\
{\scriptsize \hspace*{8mm} ($\delta t_{i+1}$ is not fixed and does not have a deterministic model).}\\
{\scriptsize \textsl{8:} \hspace*{6mm} $\bullet$ $\hat{r}_{t_i+k\tau}$: Approximation of reward for instant $t_k = t_i + k\tau$:}\\
{\scriptsize \hspace*{16mm} $\hat{r}_{t_i+k\tau} \sim f(k, \delta t_{i+1}, R(s_{t_i}), R(s_{t_{i+1}}))$}\\
{\scriptsize \textsl{9:} \hspace*{6mm} $\bullet$ $f(.)$: The approximation function. It could be of order $\geq n$.}\\ 
{\scriptsize \hspace*{8mm} $n$ depends on the design of the reward function $R(s) \in C^n$.}\\ 
{\scriptsize \hspace*{9mm} As an example, but not limited to, when $n=1$ ($R(s) \in C^1$),}\\ 
{\scriptsize \hspace*{9mm} function $f(.)$ could be equal to:}\\
{\scriptsize \hspace*{12mm} $f(.) = R(s_{t_i}) + (k + 1) \cdot \tau \cdot \left[ R(s_{t_{i+1}}) - R(s_{t_i}) \right] / \delta t_{i+1}$}\\
{\scriptsize \textsl{10:} \hspace*{4mm} Evaluate $\hat{A}^\pi(s_{t_i}, a_{t_i}) = \hat{V}^\pi_{tar}(s_{t_i}, a_{t_i}) - \hat{V}^\pi_\phi(s_{t_i}, a_{t_i})$}\\
{\scriptsize \textsl{11:} \hspace*{4mm} Evaluate $\nabla_\theta J(\theta) \approx \nabla_\theta \log \pi_\theta(a|s) \hat{A}^\pi(s, a)$}\\
{\scriptsize \textsl{12:} \hspace*{4mm} Update $\theta \leftarrow \theta + \alpha \nabla_\theta J(\theta)$}\\
{\scriptsize \textsl{13:} \hspace*{1mm} \textbf{end for}}\\
{\scriptsize \textsl{14:} \textbf{end for}}\\
\end{tabular}}\\ \hline
\end{tabular}
\end{table}
\section{Simulation framework}\label{Sim_framework}
\begin{figure*}[t]
\centering
\includegraphics[scale=0.68]{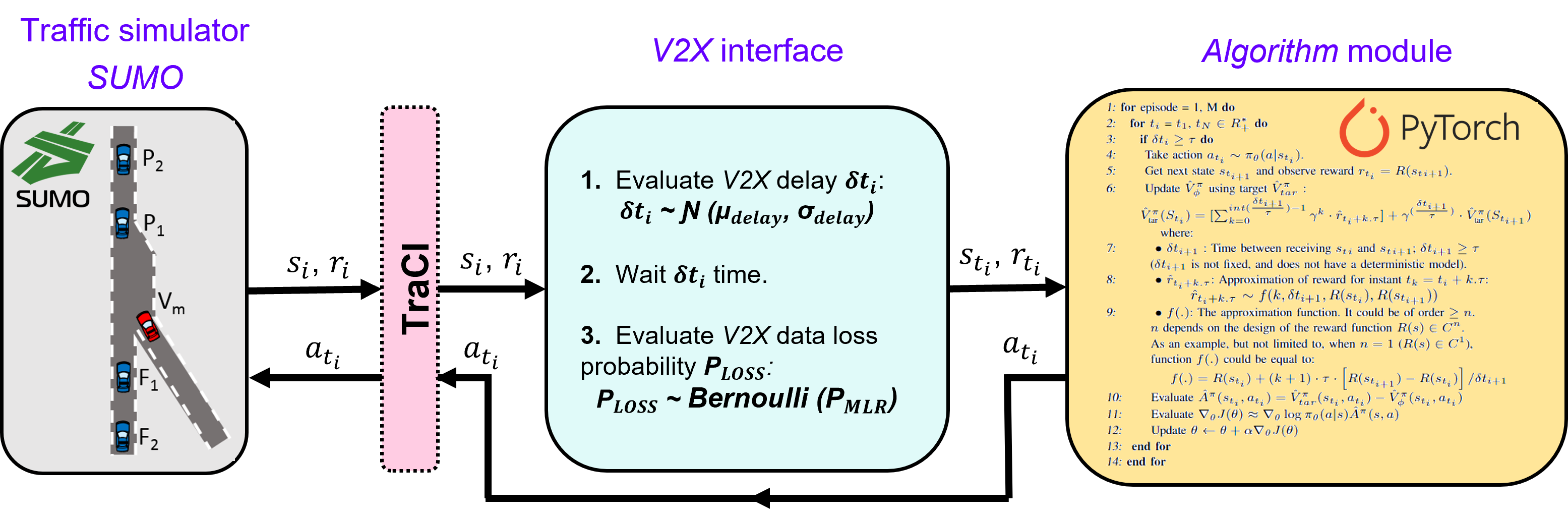}
\caption{Illustration of simulation framework. The traffic simulator provide information ($s_i$,$r_i$), periodically, to the V2X interface through \textsl{TraCI}. The V2X interface generates delays and loss and then forward information ($s_{t_i}$,$r_{t_i}$) to the algorithm accordingly. The algorithm is executed and action $a_{t_i}$ is provided to the traffic simulator through \textsl{TraCI}.}
\label{Sim_frame}
\end{figure*}

We use a simulation framework to train, test, and validate the novel algorithm (cf. Fig. \ref{Sim_frame}). The simulation framework comprises a traffic simulator environment \textsl{SUMO} (Simulation of Urban Mobility), a V2X interface to simulate V2X data delays and loss without deep diving into complex network modelling, and a module that incorporates the algorithm. The traffic simulator control the traffic and motion of vehicles and provide their states to the V2X interface through \textsl{TraCI} \cite{Traci}, which is an API that provides access to a \textsl{SUMO} traffic simulation. The V2X interface generates communication delays according to some parametric function ($\mathcal{N}(\mu_{delay},\sigma_{delay})$), and provides input state to the algorithm according to this calculated delay and the data loss probability ($P_{LOSS}$). Since estimating an accurate probability density function (PDF) for V2X communication delays is not feasible in practice because the external factors and application scenarios differ, we use the \textsl{Normal} distribution $\mathcal{N}(\mu_{delay},\sigma_{delay})$ for delays, and \textsl{Bernoulli} distribution \textsl{Bernoulli($P_{MLR}$)} for the data loss probability. For consistency, the values of these functions' parameters are selected uniformly in their specific ranges as was reported in the literature \cite{V2X_delays_reference,DSRC_LTE,DSRC_LTE_1,DSRC_LTE_2} (see Table \ref{V2X_param}). The algorithm module receives input data from the V2X interface and performs training and testing using \textsl{PyTorch} library \cite{pytorch_lib}. It then provides actions to control CAV through the same interface \textsl{TraCI}.
\begin{table}[b]
\caption{V2X parametric functions parameters' values.}
\centering
\begin{tabular}{|l|c|}
\hline
\multicolumn{1}{|c|}{\textbf{Parameter}}                                                                                & \textbf{Range}                          \\ \hline
\textbf{$\mu_{delay}$} : mean of V2X delays                  & [10 ms, 30 ms, 50 ms, 70 ms, 90 ms] \\ \hline
\begin{tabular}[c]{@{}l@{}}\textbf{$\sigma_{delay}$} : standard deviation of \\ V2X delays\end{tabular} & [23 ms]                             \\ \hline
\textbf{$P_{MLR}$}: V2X Message Loss Rate                                   & [0.1, 0.3, 0.5, 0.7, 0.9]           \\ \hline
\end{tabular}
\label{V2X_param}
\end{table}
\subsection{Use Case and Simulation Parameters}\label{UC_params}
\begin{figure}[t]
\begin{center}
\includegraphics[scale=0.18]{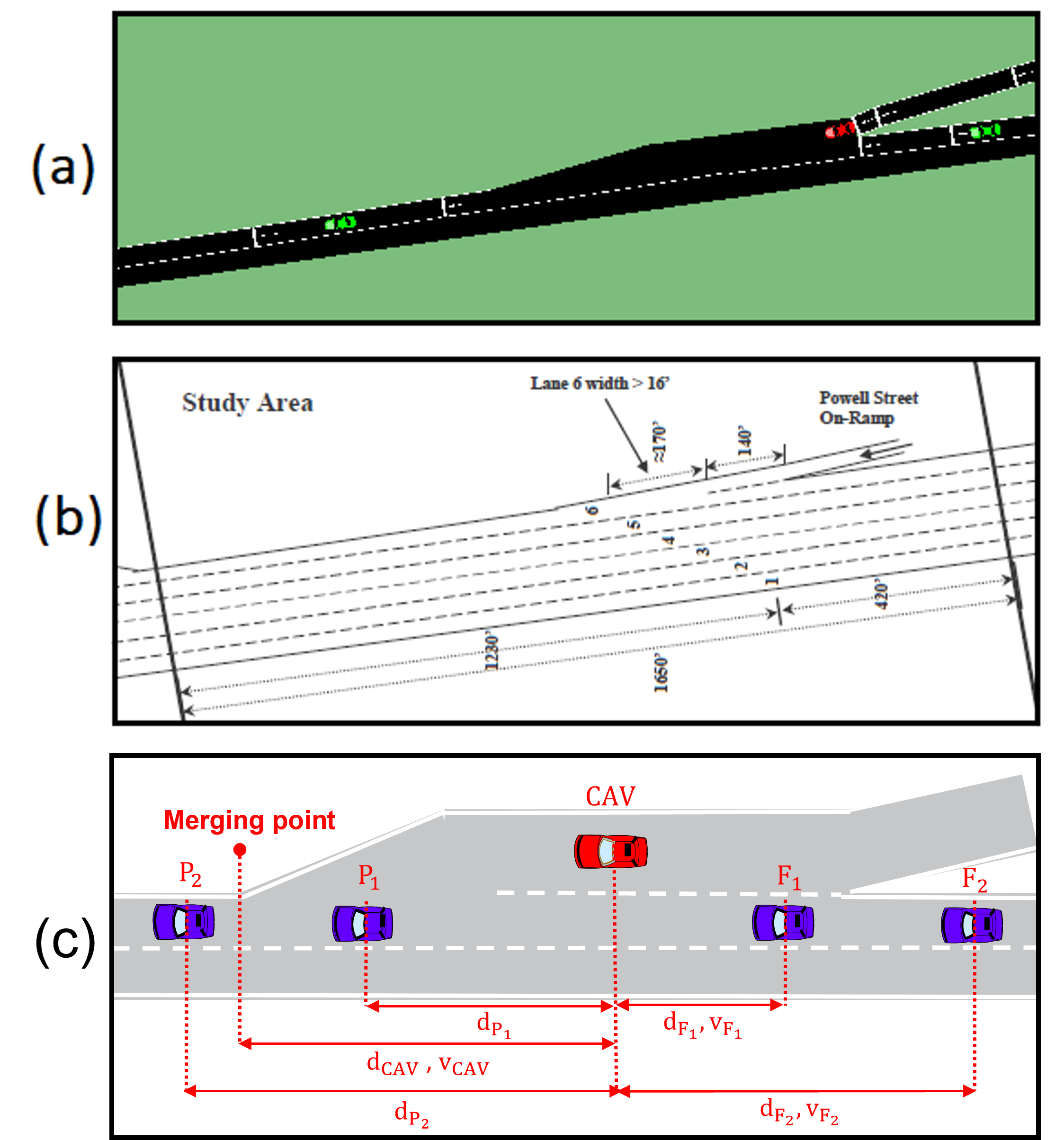}
\caption{Ramp merging (a) simulated scenario (b) real-world location geometry (c) state vector features.}
\label{Sim_scenario}
\end{center}
\vspace{-5mm}
\end{figure}

To evaluate our approach, we will train and test the novel algorithm to perform high-speed highway on-ramp merging under the simulation framework. The motivations for using this use case are:
\begin{itemize}
    \item First, highway-on ramp merging involves several complex tasks such as searching and finding appropriate gap, adjusting speed, and interacting with surrounding vehicles. This complexity enables a more faithful evaluation of the present approach compared to simpler use cases, such as vehicle following or speed maintaining.
    \item Second, highway on-ramp locations are critical zones for traffic safety. According to recent report by the National Highway Traffic Safety Administration \cite{Safety_report_1}, nearly 30,000 highway merging collisions occur each year in the USA, which represents 0.3\% of all collisions \cite{Safety_report_2}.
    \item Lastly, highway on-ramp locations are also critical zones for traffic efficiency. Indeed, 40\% of traffic congestion on the U.S. highway system is caused by recurring bottlenecks, among which, highway on-ramps are significant \cite{Efficiency_report_1}.
\end{itemize}
\indent Under the simulation framework, we will faithfully replicate a real-world highway on-ramp scenario (cf. Fig. \ref{Sim_scenario}) located on a segment of interstate 80 in Emeryville (San Francisco), California, where the traffic flow is extracted from \textsl{NGSIM} database \cite{NGSIM}. Similar to \cite{Kherroubi2022}, simulation parameters are summarized in Table \ref{Sim_param}.\\
\indent Using this simulation setup, our novel \textsl{Blind Actor-critic} algorithm is trained and tested where the state ($s_{t_i}$), action ($a_{t_i}$), and reward ($r_{t_i}$) are defined as follows:
\begin{itemize}
\item \textbf{State ($s_{t_i}$)} is defined as a vector:\\$s_{t_i}=<d_{CAV},~v_{CAV},~d_{P_1},~d_{F_1},~v_{F_1},~d_{P_2},~d_{F_2},~v_{F_2}>$ where $d_{CAV}$ (resp. $v_{CAV}$) is the distance to the merging point (resp. speed) of CAV, while $d_{K_i}$ (resp. $v_{K_i}$) is the relative distance (resp. speed) between vehicle \textsl{$K_i$} and CAV, as illustrated in Fig. \ref{Sim_scenario}.(c). This vector comprises only the most significant features \cite{Kherroubi2022}.
\item \textbf{Action ($a_{t_i}$)} is the longitudinal kinetic acceleration of CAV.
\end{itemize}
\begin{center}
\begin{table}[H]
\caption{Simulation's parameters values \cite{Kherroubi2022}.}
\centering
\begin{tabular}{|c|c|} 
\cline{1-2}
\textbf{Parameter}                 & \textbf{Value}    \\
\cline{1-2}
Main lane speed limit \textsl{($v_{limit}$)} &   \textsl{33 m/s} \\ 
\cline{1-2}
Main lane speed range       &   \textsl{[22 m/s, 34 m/s]} \\ 
\cline{1-2}
Main lane traffic flow &   \begin{tabular}[c]{@{}c@{}}{$\mathcal{N}(\mu\,,\sigma)$} \\ {$\mu=1\,vehicle\,per\,\textsl{3.25 seconds}$}\\{$\sigma=\textsl{0.1}$}\end{tabular}\\ 
\cline{1-2}
Main lane acceleration range & \textsl{[-5 m/$s^{2}$, +3 m/$s^{2}$]}\\ 
\cline{1-2}
Maximum emergency deceleration & \textsl{-9 m/$s^{2}$}\\ 
\cline{1-2}
\begin{tabular}[c]{@{}c@{}}Main lane driver's cooperation\\ level \textsl{(C)} \end{tabular}&   \begin{tabular}[c]{@{}l@{}}{\hspace{11mm}\textsl{C $\in$ [$C_{min}$; 1]}} \\ {{\scriptsize\textsl{C=1:}} most cooperative driver}\\{{\scriptsize\textsl{C=$C_{min}$:}} least cooperative driver}\end{tabular}\\ 
\cline{1-2}
\end{tabular}
\label{Sim_param}
\end{table}
\end{center}
\begin{itemize}
\item \textbf{Reward ($r_{t_i}$)} is defined as follow:
\[r_{t_i}=\left\{
                \begin{array}{ll}
                  -\alpha |e^{-(\frac{d_{P_{1}}}{100})} - e^{-(\frac{d_{F_{1}}}{100})}|~~~~(*)\\
                  +1~~~~~~~~~~~~~~~~~~~~~~~~~~~~(**)\\
                  -1~~~~~~~~~~~~~~~~~~~~~~~~~~~~(***)
                \end{array}
              \right.
  \]
where:\\
\textbf{(*)} This function aims to maximize safety distance with the preceding and following vehicles when CAV is in the merge zone. $\alpha$ is a hyper-parameter that will be automatically tuned during training.\\
\textbf{(**)} CAV completed the merging successfully.\\
\textbf{(***)} CAV performed a collision or a stop.
\end{itemize}
\begin{center}
\begin{table}[t]
\caption{Hyper-parameters values}
\centering
\begin{tabular}{|c|c|} 
\cline{1-2}
\textbf{Hyper-parameter}       & \textbf{Value}   \\ 
\cline{1-2}
Reward discount factor                    & 0.98    \\ 
\cline{1-2}
Actor learning rate                    & 0.0001    \\ 
\cline{1-2}
Critic learning rate                   & 0.001    \\ 
\cline{1-2}
Target network update coefficient       & 0.001   \\
\cline{1-2}
Experience replay memory size                    & 400000    \\ 
\cline{1-2}
Mini-batch size                    & 64    \\ 
\cline{1-2}
\textsl{Ornstein-Uhlenbeck} $\sigma$                    & 0.4    \\ 
\cline{1-2}
\textsl{Ornstein-Uhlenbeck} $\Theta$                   & 0.2    \\
\cline{1-2}
\end{tabular}
\label{hyp_params}
\vspace{-10mm}
\end{table}
\end{center}
\section{Results and discussions}\label{Res_disc}
The \textsl{Blind Actor-critic} was trained and tested to perform highway on-ramp merging under the simulation framework. The performances are then compared to benchmark algorithms in order to validate our novel approach. The training was performed using \textsl{Automatic hyper-parameters tuning} to preserve consistency, and ensure fairness and comparability between approaches. For time and cost effectiveness, only the following hyper-parameters were automatically tuned: the reward factor $\alpha$, the number of training steps, and the fictive factor $\tau$. The fixed hyper-parameters values are summarized in Table \ref{hyp_params}.\\
\indent In the following, we will present and discuss the training and testing performance.
\subsection{Training Performance}\label{train_perf}
Three metrics are used to compare the training efficiency: \textsl{the residual variance of the value function}, \textsl{the average normalized training rewards}, and \textsl{the cumulative approximation errors}.
\subsubsection{\textbf{Evaluation of Residual Variance of Value Function}}
\begin{figure}[t]
\begin{center}
\includegraphics[scale=0.26]{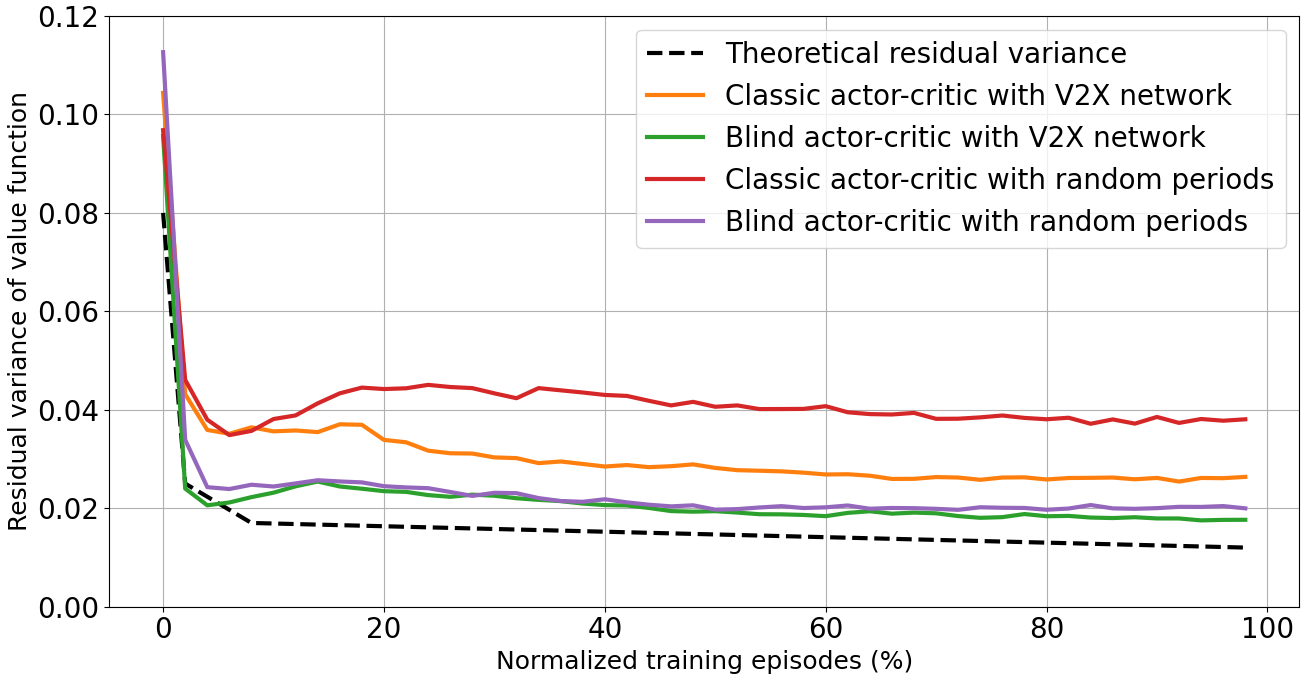}
\caption{Average residual variance of value function during training.}
\label{Resid_var}
\end{center}
\vspace{-6mm}
\end{figure}

The residual variance of the value function is defined by equation (\ref{eq_resid}):
\begin{equation}
    Residual~variance = \frac{Var(\hat{V}^\pi_{tar}-\hat{V}^\pi_{\phi})}{Var(\hat{V}^\pi_{tar})},
\label{eq_resid}
\end{equation}
where \textsl{Var(.)} denotes the variance. The residual variance quantifies how well the trained critic network $\hat{V}^\pi_\phi$ has learnt the true values of the empirical target critic $\hat{V}^\pi_{tar}$. A high value of residual variance at the end of training indicates that the value estimator $\hat{V}^\pi_\phi$ fails to fit the true values $V^\pi$ and, hence, it fails to optimize the objective rewards \cite{Residual_var}. Consequently, high residual variance can negatively impact the learning and performance of the actor $\pi_\theta$.\\ 
We evaluated the residual variance of value function for the \textsl{Blind Actor-critic} and compare it with that of the classic Actor-critic, under different delays conditions. The results are presented in Fig. \ref{Resid_var}. For illustrative purpose, the theoretical curve of residual variance is also shown in Fig. \ref{Resid_var}, using a black dashed line, where the curve drops drastically at the beginning of training, then it decreases gradually toward zero. When performing training with V2X network data $(\mu_{delay}=50~ms,\sigma_{delay}=23~ms, P_{MLR}=0.7)$, the \textsl{Blind Actor-critic} (green curve in Fig. \ref{Resid_var}) reaches lower residual variance values at the end of training compared to the classic algorithm (orange curve in Fig. \ref{Resid_var}). To confirm these results, we increased the temporal aperiodicity level of data by generating random and uniform delays in the range of [0 ms, 1200 ms]. From \hbox{Fig. \ref{Resid_var}}, the \textsl{Blind Actor-critic} (purple curve) keeps lower values of residual variance at the end of training in contrast to the classic approach (red curve) whose residual variance increased significantly. For both cases (V2X network data and random uniform delays), the average residual variance is smoother when training the \textsl{Blind Actor-critic} algorithm compared to the classic algorithm where an apparent oscillation occurs around the elbow point of the theoretical curve (around 0\%-20\% of training episodes).\\ 
According to these results, the novel algorithm preserves the accuracy and reliability of the value function estimation even when trained using delayed and lost data.
\subsubsection{\textbf{Evaluation of Average Training Reward}} 
\begin{figure}[b]
\begin{center}
\includegraphics[scale=0.322]{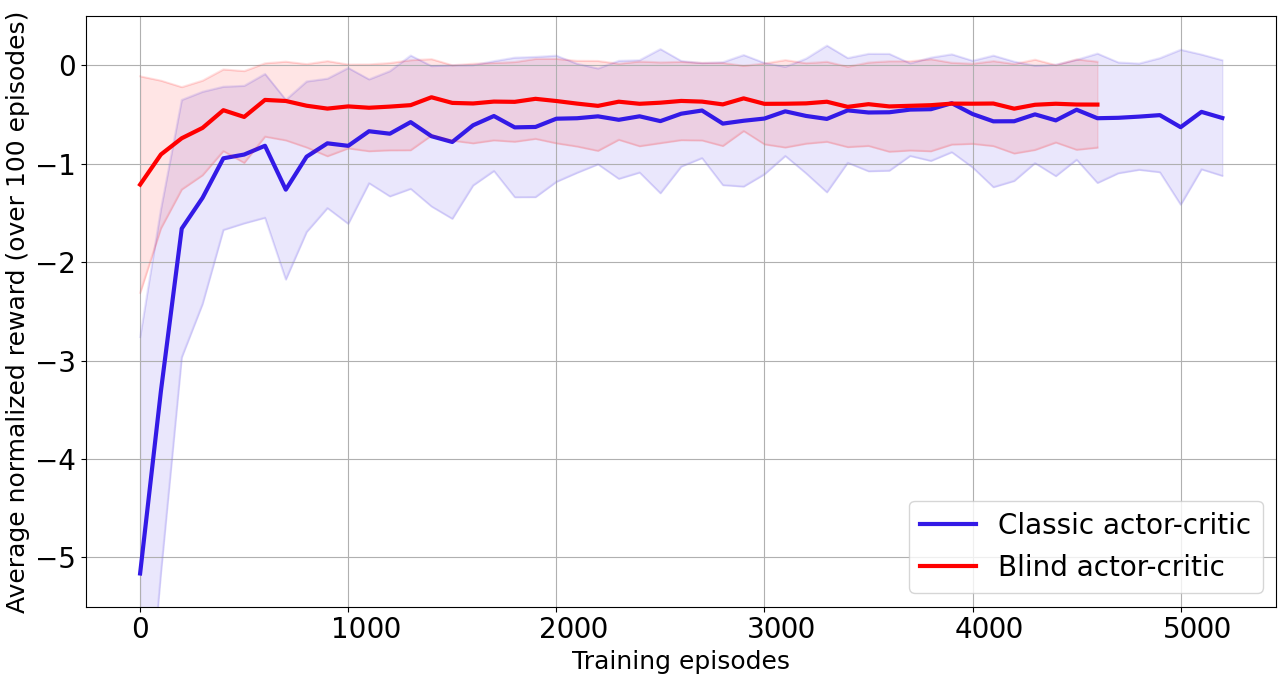}
\caption{Average normalized reward over 100 episodes when training with V2X network data.}
\label{Avrg_reward}
\end{center}
\end{figure}

We evaluated the average training reward of our \textsl{Blind Actor-critic} and compare it to that of a classic actor-critic, using V2X network data $(\mu_{delay}= 50~ms,\sigma_{delay}= 23~ms, P_{MLR}=0.7)$. For consistency and fairness, the reward is further normalized by the hyper-parameter $\alpha$. The results are presented in Fig. \ref{Avrg_reward}. As shown in this figure, the \textsl{Blind Actor-critic} converges faster compared to the classic algorithm. It also achieves higher average reward values early in the training phase (around 0\%-20\% of training episodes), which confirms the findings about the residual variance of the value function. The average rewards of the \textsl{Blind Actor-critic} has smoother curve and lower variance (see dashed zones in Fig. \ref{Avrg_reward}). This could be explained by the introduced fictive sampling period $\tau$ and reward approximation mechanism.
\\
\begin{figure}[t]
\centering
\subfloat[]{\includegraphics[scale=0.44]{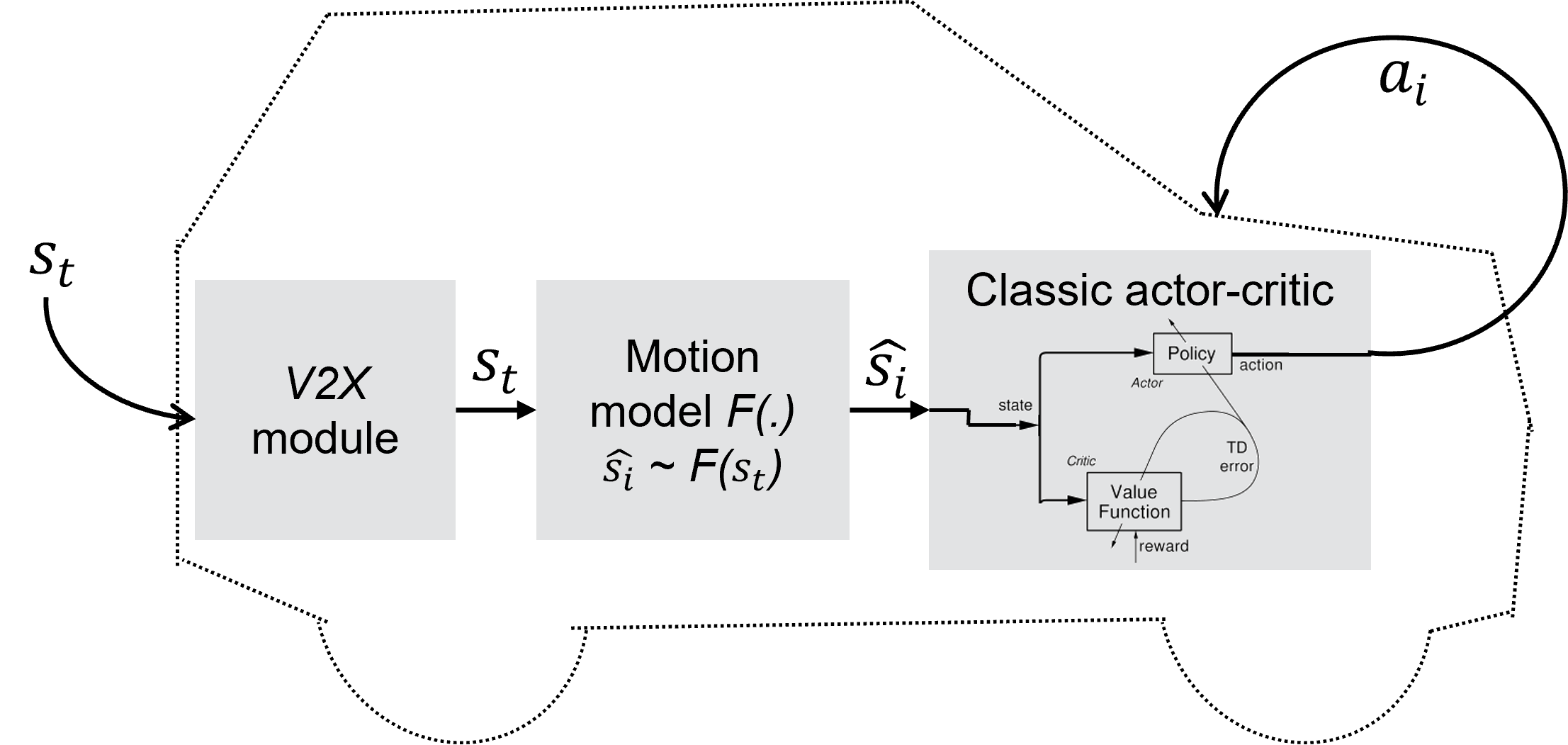}}
\hfil
\subfloat[]{\includegraphics[scale=0.27]{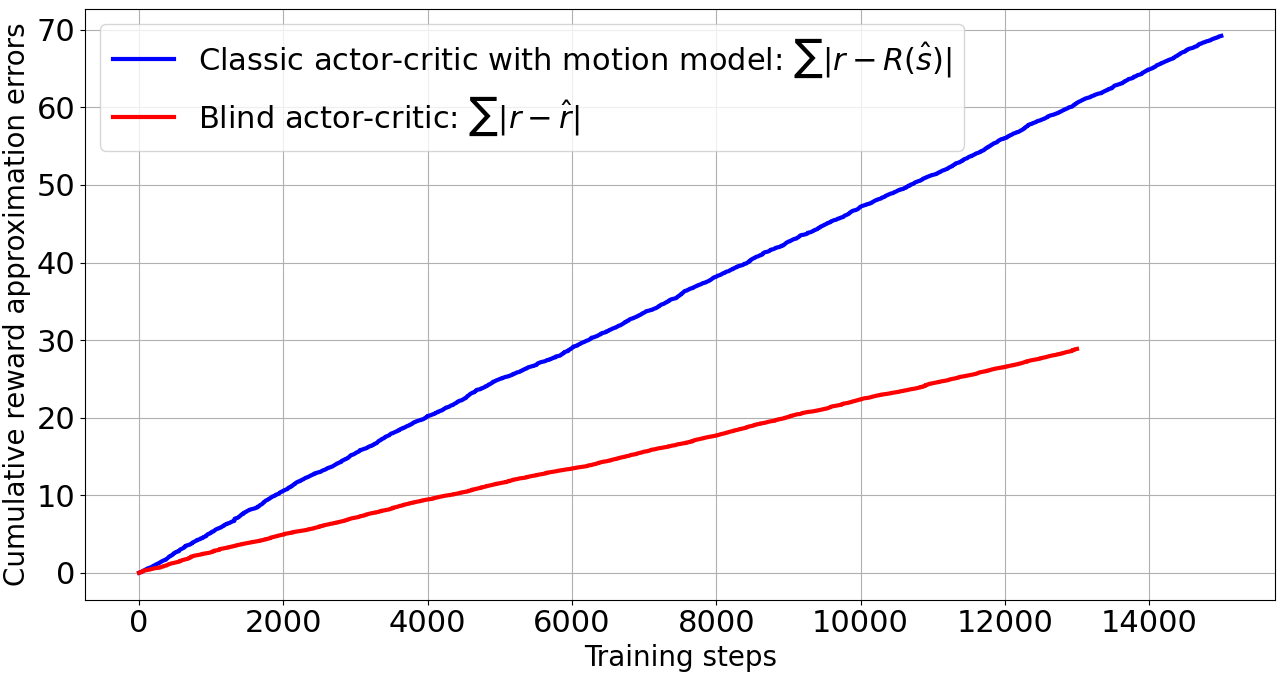}}
\hfil
\subfloat[]{\includegraphics[scale=0.27]{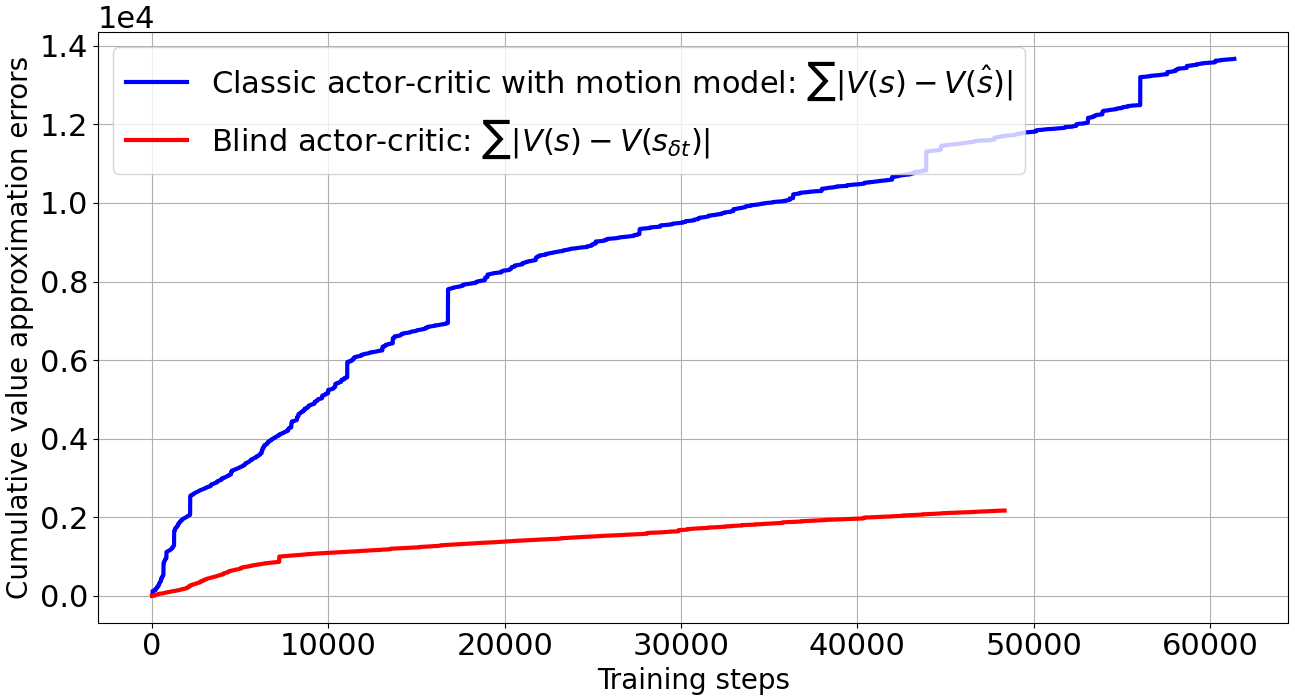}}
\caption{Approximation errors analysis. (a) Illustration of benchmark approach with approximate motion model. (b) Cumulative reward approximation errors. (c) Cumulative value approximation errors.}
\label{Approx_error}
\vspace{-5mm}
\end{figure}
\subsubsection{\textbf{Evaluation of Cumulative approximation Errors}} In order to guarantee the robustness of reward approximations that were introduced, approximation errors during training were evaluated and compared to those of a benchmark approach that use classic Actor-critic algorithm with state estimation, as illustrated in Fig. \ref{Approx_error}.(a). This latest benchmark approach use classic Actor-critic with fixed sampling periods where the missing values of states and rewards, due to V2X network delays and loss, are estimated using an approximate motion model \cite{motion_approx_1, motion_approx_2}, as illustrated in Fig. \ref{Approx_error}.(a). Results show that the \textsl{Blind Actor-critic} induces lower approximation errors for rewards (cf. Fig. \ref{Approx_error}.(b)). This is due to the fact that reward approximation is performed \textsl{scalarly} using available ground truth values of rewards, in contrast to the benchmark approach with motion model, where the approximation is performed on the multi-dimensional state vector. The approximation errors of the value function are also considerably lower when training the \textsl{Blind Actor-critic} (cf. Fig. \ref{Approx_error}.(c)) because the learning of the critic is performed using \textsl{Temporal-Difference} with the ground truth of available states and a modulated discount factor (i.e. no approximation of the state vector is performed).
\subsection{Testing Performance}\label{test_perf}
\begin{figure*}[h]
\centering
\includegraphics[scale=0.156]{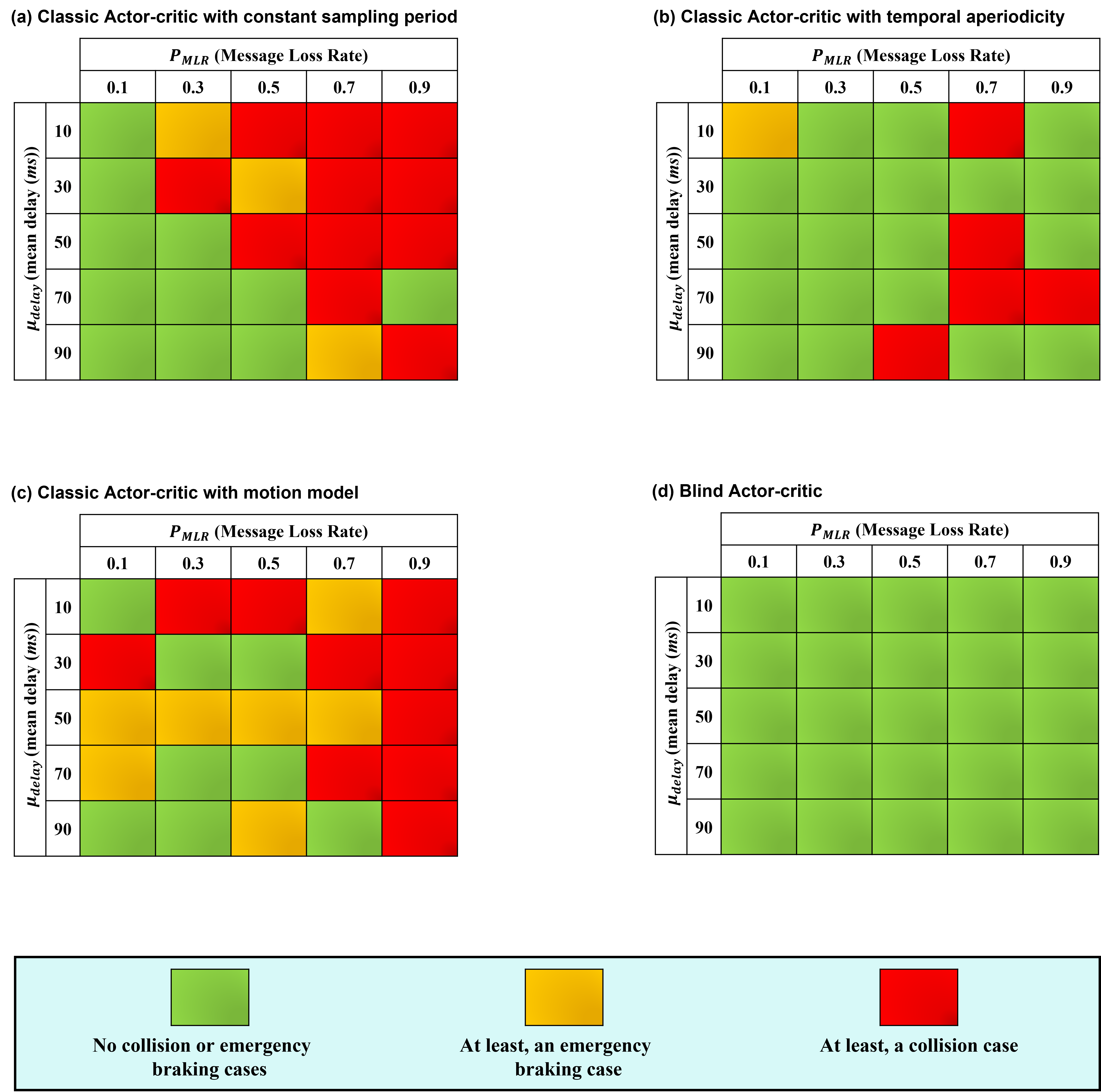}
\caption{Testing performance for different V2X network reliability levels.}
\label{testing_perf}

\bigskip 

\captionof{table}{Performance metrics results}
\begin{tabular}{|c|c|c|c|c|}
\hline
\small{\textbf{Performance metric}}&  
\begin{tabular}[c]{@{}c@{}} \small{\textbf{Actor-critic with}}\\ \small{\textbf{constant sampling}}\end{tabular}
& 
\begin{tabular}[c]{@{}c@{}} \small{\textbf{Actor-critic with}}\\ \small{\textbf{temporal aperiodicity}}\end{tabular} 
& 
\begin{tabular}[c]{@{}c@{}} \small{\textbf{Actor-critic with}}\\ \small{\textbf{motion model}}\end{tabular} 
& \small{\textbf{\textsl{Blind Actor-critic}}}
\\ \hline

\small{\textbf{Average safety distance \textsl{(m)}}}                                                          & \small{53.31}    & \small{52.86}    & \small{53.21}     & \small{53.81}                                                       \\ \hline
\small{\textbf{Number of collisions}}  & \small{15}    & \small{7}                                                         & \small{15}     & \small{0}                                                        \\ \hline
\small{\textbf{Number of emergency brakings}}   & \small{23}    & \small{7}         & \small{27}     & \small{0}                                                        \\ \hline
\small{\textbf{Average speed \textsl{(km/h)}}}  & \small{109} & \small{109} & \small{113} & \small{114}                                               \\ \hline
\end{tabular}
\label{perf_table}
\end{figure*}

To validate the robustness of the \textsl{Blind Actor-critic} regarding delays and data loss in V2X networks, we tested and compared its performance to those of benchmark methods. For consistency and fairness, hard testing conditions were used, as follows:
\begin{itemize}
    \item V2X network reliability level is gradually degraded for various values of delays and data loss:
\end{itemize}
$\mu_{delay} \times P_{MLR} \rightarrow [10, 30, 50, 70, 90] \times [0.1, 0.3, 0.5, 0.7, 0.9]$
\begin{itemize}
    \item For each V2X network reliability level, 10000 highway on-ramp merging episodes were tested under the simulation framework. This high number of testing episodes allows to evaluate and compare the asymptotic performance of each approach faithfully.
    \item For each V2X network reliability level, safety (number of collisions and average safety distance) and efficiency (average speed) performance metrics are evaluated over the 10000 merging episodes.
\end{itemize}
Four approaches are tested and their performances are compared:
\begin{enumerate}
    \item Classic Actor-critic that is trained at constant sampling period (i.e. without temporal aperiodicity).
    \item Classic Actor-critic that is trained with V2X network data $(\mu_{delay}= 50~ms,\sigma_{delay}= 23~ms, P_{MLR}=0.7)$.
    \item Classic Actor-critic with approximate motion model, that is trained with V2X network data $(\mu_{delay}= 50~ms,\sigma_{delay}= 23~ms, P_{MLR}=0.7)$.
    \item \textsl{Blind Actor-critic} that is trained with V2X network data $(\mu_{delay}= 50~ms,\sigma_{delay}= 23~ms, P_{MLR}=0.7)$.
\end{enumerate}
The testing results are summarized in Fig. \ref{testing_perf}. As was previously stated in section \ref{SOTA}, Actor-critic performance deteriorates under low V2X network reliability level (see Fig. \ref{testing_perf}.(a)), even when trained with delayed and lost data (see Fig. \ref{testing_perf}.(b)) or using approximate motion model (see Fig. \ref{testing_perf}.(c)). When the V2X communication link is degraded (i.e. higher delays and Message Loss Rate), collision and/or emergency barking cases may occur. From these results, we noticed that performances are more sensitive to the Message Loss Rate ($P_{MLR}\ge 0.5$) compared to the mean delay ($\mu_{delay}$). In contrast to the benchmark approaches, our approach maintains a zero rate for collisions and emergency brakings (see Fig. \ref{testing_perf}.(d)) even when V2X communication link is degraded. This confirm the robustness of the \textsl{Blind Actor-critic} algorithm.\\
The performance metrics of each tested approach are also summarized in Table \ref{perf_table}. Results show that the \textsl{Blind Actor-critic} guarantees a higher safety distance compared to remaining approaches. It also prevents collisions and emergency braking cases even at lower V2X network reliability levels. Concerning traffic efficiency, the average speed of merging is also slightly higher when using the \textsl{Blind Actor-critic}, which means that the traffic flow is more efficient.
\newpage
\section{Conclusion}\label{conc}
In this paper, we presented a novel approach, \textsl{Blind Actor-critic}, for controlling CAV to compensate for delays and data loss in V2X network. Indeed, simulation results have demonstrated that performance of classic AI algorithms deteriorate when data is received under low V2X network reliability level. To overcome this challenge, the novel approach is built on top of reinforcement learning Actor-critic architecture. The robustness of the proposed approach is due to three main components that were introduced: a fictive constant sampling period, a combination between \textsl{Temporal-Difference} and \textsl{Monte Carlo} learning, and a numerical approximation of unavailable reward values. Results have shown that this novel approach improves the training with lower residual variance of the value function, and higher and smother rewards, while keeping low approximation errors. These improvements in the training phase are translated into improvements in the testing performance. Indeed, our approach maintains zero rate of collisions and high average speed, even under distorted V2X communication. In other words, the novel \textsl{Blind Actor-critic} preserves the control robustness of CAV while maintaining high driving safety and efficiency.\\
\indent As perspectives, this approach could be scaled to different simulated scenarios, such as driving at intersections. For that, more advanced simulation environment can be used to include deeper V2X communication models and vehicles’ dynamic. The next steps toward real-world deployment is to experiment the method in practice using either small scale testbeds or dedicated experimentation areas. Future researches should also extend the applications to other domains beyond autonomous driving, such as remote control of robots, unmanned aerial vehicles (UAV), underwater vehicles, etc. Questions about the use of the proposed approach to control systems with delays are still awaiting to be asked and answered. 


\bibliographystyle{IEEETran}
\bibliography{Blind_Actor_Critic} 

\begin{thebibliography}{10}
\providecommand{\url}[1]{#1}
\csname url@samestyle\endcsname
\providecommand{\newblock}{\relax}
\providecommand{\bibinfo}[2]{#2}
\providecommand{\BIBentrySTDinterwordspacing}{\spaceskip=0pt\relax}
\providecommand{\BIBentryALTinterwordstretchfactor}{4}
\providecommand{\BIBentryALTinterwordspacing}{\spaceskip=\fontdimen2\font plus
\BIBentryALTinterwordstretchfactor\fontdimen3\font minus \fontdimen4\font\relax}
\providecommand{\BIBforeignlanguage}[2]{{%
\expandafter\ifx\csname l@#1\endcsname\relax
\typeout{** WARNING: IEEEtran.bst: No hyphenation pattern has been}%
\typeout{** loaded for the language `#1'. Using the pattern for}%
\typeout{** the default language instead.}%
\else
\language=\csname l@#1\endcsname
\fi
#2}}
\providecommand{\BIBdecl}{\relax}
\BIBdecl

\bibitem{CV_crash_rate_1}
J.~B. Kenney, ``Dedicated short-range communications (dsrc) standards in the united states,'' \emph{Proceedings of the IEEE}, vol.~99, pp. 1162--1182, 2011.

\bibitem{CV_crash_rate_2}
M.~Noor-A-Rahim, G.~G.~N. Ali, Y.~L. Guan, B.~Ayalew, P.~H.~J. Chong, and D.~Pesch, ``Broadcast performance analysis and improvements of the lte-v2v autonomous mode at road intersection,'' \emph{IEEE Transactions on Vehicular Technology}, vol.~68, pp. 9359--9369, 10 2019.

\bibitem{ETSI_CAM}
ETSI, ``Intelligent transport systems (its); vehicular communications; basic set of applications; part 2: Specification of cooperative awareness basic service,'' 4 2019.

\bibitem{SAE_CAM}
SAE, ``V2x communications message set dictionary j2735\_202211,'' 11 2022.

\bibitem{ETSI_CPM}
ETSI, ``Intelligent transport systems (its); vehicular communications; basic set of applications; analysis of the collective perception service (cps); release 2,'' 12 2019.

\bibitem{CPM_accuracy_1}
\BIBentryALTinterwordspacing
X.~Chen, J.~Ji, and Y.~Wang, ``Robust cooperative multi-vehicle tracking with inaccurate self-localization based on on-board sensors and inter-vehicle communication,'' \emph{Sensors 2020, Vol. 20, Page 3212}, vol.~20, p. 3212, 6 2020. [Online]. Available: \url{https://www.mdpi.com/1424-8220/20/11/3212/htm https://www.mdpi.com/1424-8220/20/11/3212}
\BIBentrySTDinterwordspacing

\bibitem{CPM_accuracy_2}
A.~Miller, K.~Rim, P.~Chopra, P.~Kelkar, and M.~Likhachev, ``Cooperative perception and localization for cooperative driving,'' \emph{Proceedings - IEEE International Conference on Robotics and Automation}, pp. 1256--1262, 5 2020.

\bibitem{CPM_accuracy_3}
Q.~Chen, S.~Tang, Q.~Yang, and S.~Fu, ``Cooper: Cooperative perception for connected autonomous vehicles based on 3d point clouds,'' \emph{Proceedings - International Conference on Distributed Computing Systems}, vol. 2019-July, pp. 514--524, 7 2019.

\bibitem{DRL}
T.~Lillicrap, J.~J. Hunt, A.~Pritzel, N.~Heess, T.~Erez, Y.~Tassa, D.~Silver, and D.~Wierstra, ``Continuous control with deep reinforcement learning,'' \emph{CoRR}, 2015.

\bibitem{Ping_Wang}
\BIBentryALTinterwordspacing
P.~Wang and C.-Y. Chan, ``Autonomous ramp merge maneuver based on reinforcement learning with continuous action space,'' \emph{CoRR}, vol. abs/1803.09203, 2018. [Online]. Available: \url{http://arxiv.org/abs/1803.09203}
\BIBentrySTDinterwordspacing

\bibitem{Kherroubi2022}
Z.~E.~A. Kherroubi, S.~Aknine, and R.~Bacha, ``Novel decision-making strategy for connected and autonomous vehicles in highway on-ramp merging,'' \emph{IEEE Transactions on Intelligent Transportation Systems}, vol.~23, pp. 12\,490--12\,502, 8 2022.

\bibitem{Sutton}
R.~S. Sutton and A.~G. Barto, \emph{Reinforcement Learning: An Introduction}, 2nd~ed.\hskip 1em plus 0.5em minus 0.4em\relax The MIT Press, 2015.

\bibitem{Delayed_AC}
Y.~Bouteiller, S.~Ramstedt, G.~Beltrame, C.~Pal, and J.~Binas, ``Reinforcement learning with random delays,'' 2021.

\bibitem{Mujoco}
\BIBentryALTinterwordspacing
DeepMind. Mujoco: Advanced physics simulation. [Online]. Available: \url{https://mujoco.org/}
\BIBentrySTDinterwordspacing

\bibitem{SAE_CPM}
SAE, ``V2x vehicular applications technical committee. cooperative perception system j2945/8,'' 9 2018.

\bibitem{ETSI_V2X_app}
ETSI, ``Etsi tr 102 638; intelligent transport systems (its); vehicular communications; basic set of applications; definitions,'' 6 2009.

\bibitem{V2X_delays_reference}
Y.~Fang, H.~Min, X.~Wu, W.~Wang, X.~Zhao, and G.~Mao, ``On-ramp merging strategies of connected and automated vehicles considering communication delay,'' \emph{IEEE Transactions on Intelligent Transportation Systems}, pp. 1--15, 2022.

\bibitem{DSRC_LTE}
M.~Shi, C.~Lu, Y.~Zhang, and D.~Yao, ``Dsrc and lte-v communication performance evaluation and improvement based on typical v2x application at intersection,'' \emph{Proceedings - 2017 Chinese Automation Congress, CAC 2017}, vol. 2017-January, pp. 556--561, 12 2017.

\bibitem{Wireless_book}
A.~F. Molisch, \emph{Wireless Communications}, 2nd~ed.\hskip 1em plus 0.5em minus 0.4em\relax John Wiley and Sons Ltd, 2011.

\bibitem{CCA_V2X}
S.~Biswas, R.~Tatchikou, and F.~Dion, ``Vehicle-to-vehicle wireless communication protocols for enhancing highway traffic safety,'' \emph{IEEE Communications Magazine}, vol.~44, pp. 74--82, 1 2006.

\bibitem{HU2019506}
\BIBentryALTinterwordspacing
S.~G. Hu, H.~Y. Wen, L.~Xu, and H.~Fu, ``Stability of platoon of adaptive cruise control vehicles with time delay,'' \emph{Transportation Letters}, vol.~11, no.~9, pp. 506--515, 2019. [Online]. Available: \url{https://www.sciencedirect.com/science/article/pii/S1942786722001539}
\BIBentrySTDinterwordspacing

\bibitem{WANG2018276}
\BIBentryALTinterwordspacing
M.~Wang, ``Infrastructure assisted adaptive driving to stabilise heterogeneous vehicle strings,'' \emph{Transportation Research Part C: Emerging Technologies}, vol.~91, pp. 276--295, 2018. [Online]. Available: \url{https://www.sciencedirect.com/science/article/pii/S0968090X18304674}
\BIBentrySTDinterwordspacing

\bibitem{platoon_CACC}
S.~Hasan, M.~A. Al~Ahad, I.~Sljivo, A.~Balador, S.~Girs, and E.~Lisova, ``A fault-tolerant controller manager for platooning simulation,'' in \emph{2019 IEEE International Conference on Connected Vehicles and Expo (ICCVE)}, 2019, pp. 1--6.

\bibitem{motion_estimation}
Z.~Wang, K.~Han, and P.~Tiwari, ``Motion estimation of connected and automated vehicles under communication delay and packet loss of v2x communications,'' 01 2021.

\bibitem{Patent_1}
K.~Chi, Panjju, Y.~Gouang, and W.~Jingao, ``System delay compensation control method for autonomous vehicles,'' 2020, patent number: KR102126621B1, https://patents.google.com/patent/KR102126621B1/en.

\bibitem{patent_2}
Kim, B.~Yu, T.~Hayashi, and S.~Shiraishi, ``Selective remote control of vehicle adas function,'' 2020, patent number: JP6729739B2, https://patents.google.com/patent/JP6729739B2/en.

\bibitem{Traffic_control}
I.~Finkelberg, T.~Petrov, A.~Gal-Tzur, N.~Zarkhin, P.~Pocta, T.~Kovacikova, L.~Buzna, M.~Dado, and T.~Toledo, ``The effects of vehicle-to-infrastructure communication reliability on performance of signalized intersection traffic control,'' \emph{IEEE Transactions on Intelligent Transportation Systems}, vol.~23, pp. 15\,450--15\,461, 9 2022.

\bibitem{RL_challenges}
G.~Dulac-Arnold, D.~Mankowitz, and T.~Hester, ``Challenges of real-world reinforcement learning,'' 2019.

\bibitem{Credit_assign}
M.~Minsky, ``Steps toward artificial intelligence,'' \emph{Proceedings of the IRE}, vol.~49, no.~1, pp. 8--30, 1961.

\bibitem{Blind_RL}
\BIBentryALTinterwordspacing
M.~Agarwal and V.~Aggarwal, ``Blind decision making: Reinforcement learning with delayed observations,'' \emph{Pattern Recognition Letters}, vol. 150, pp. 176--182, 2021. [Online]. Available: \url{https://www.sciencedirect.com/science/article/pii/S0167865521002282}
\BIBentrySTDinterwordspacing

\bibitem{Delayed_DQN}
\BIBentryALTinterwordspacing
K.~Du, L.~Wang, Y.~Liu, H.~Niu, S.~Huang, and X.~Wen, ``Random-delay-corrected deep reinforcement learning framework for real-world online closed-loop network automation,'' \emph{Applied Sciences}, vol.~12, no.~23, 2022. [Online]. Available: \url{https://www.mdpi.com/2076-3417/12/23/12297}
\BIBentrySTDinterwordspacing

\bibitem{Traci}
\BIBentryALTinterwordspacing
G.~A. Center. Traci - sumo documentation. Accessed: 18-09-2021. [Online]. Available: \url{https://sumo.dlr.de/docs/TraCI.html}
\BIBentrySTDinterwordspacing

\bibitem{DSRC_LTE_1}
\BIBentryALTinterwordspacing
T.~Petrov, L.~Sevcik, P.~Pocta, and M.~Dado, ``A performance benchmark for dedicated short-range communications and lte-based cellular-v2x in the context of vehicle-to-infrastructure communication and urban scenarios,'' \emph{Sensors}, vol.~21, no.~15, 2021. [Online]. Available: \url{https://www.mdpi.com/1424-8220/21/15/5095}
\BIBentrySTDinterwordspacing

\bibitem{DSRC_LTE_2}
R.~Protzmann, ``V2x communication in heterogeneous networks,'' PhD thesis, Elektrotechnik und Informatik, Technische Universit\"{a}t Berlin, Berlin, Germany, March 2018.

\bibitem{pytorch_lib}
\BIBentryALTinterwordspacing
Pytorch. Accessed: 18-09-2021. [Online]. Available: \url{https://pytorch.org/}
\BIBentrySTDinterwordspacing

\bibitem{Safety_report_1}
\BIBentryALTinterwordspacing
N.~H. T.~S. Administration and U.~D. of~Transportation, ``2016 data: Summary of motor vehicle crashes,'' Tech. Rep., 2016. [Online]. Available: \url{https://crashstats.nhtsa.dot.gov/Api/Public/ViewPublication/812501}
\BIBentrySTDinterwordspacing

\bibitem{Safety_report_2}
NHTSA, ``Analysis of lane change crashes,'' U.S. Department of Transportation, Tech. Rep., 2003.

\bibitem{Efficiency_report_1}
\BIBentryALTinterwordspacing
FHWA, ``Traffic congestion and reliability trends and advanced strategies for congestion mitigation,'' US Department of Transportation, Tech. Rep., 2005. [Online]. Available: \url{www.camsys.com}
\BIBentrySTDinterwordspacing

\bibitem{NGSIM}
\BIBentryALTinterwordspacing
U.~S.~D. of~Transportation. (2022) Next generation simulation (ngsim) vehicle trajectories and supporting data | department of transportation - data portal. [Online]. Available: \url{https://data.transportation.gov/Automobiles/Next-Generation-Simulation-NGSIM-Vehicle-Trajector/8ect-6jqj}
\BIBentrySTDinterwordspacing

\bibitem{Residual_var}
F.-B. Yannis, O.~Reda, M.~Odalric-Ambrym, and P.~Philippe, ``Learning value functions in deep policy gradients using residual variance,'' in \emph{International Conference on Learning Representations 2021}, 2021.

\bibitem{motion_approx_1}
\BIBentryALTinterwordspacing
C.~Tan, N.~Zhou, F.~Wang, K.~Tang, and Y.~Ji, ``Real-time prediction of vehicle trajectories for proactively identifying risky driving behaviors at high-speed intersections,'' \emph{Transportation Research Record}, vol. 2672, no.~38, pp. 233--244, 2018. [Online]. Available: \url{https://doi.org/10.1177/0361198118797211}
\BIBentrySTDinterwordspacing

\bibitem{motion_approx_2}
\BIBentryALTinterwordspacing
R.~Izquierdo, {\'A}.~Quintanar, D.~F. Llorca, I.~G. Daza, N.~Hern{\'a}ndez, I.~Parra, and M.~{\'A}. Sotelo, ``Vehicle trajectory prediction on highways using bird eye view representations and deep learning,'' \emph{Applied Intelligence}, vol.~53, no.~7, pp. 8370--8388, Apr 2023. [Online]. Available: \url{https://doi.org/10.1007/s10489-022-03961-y}
\BIBentrySTDinterwordspacing

\end{thebibliography}


\vspace{11pt}

\begin{IEEEbiography}[{\includegraphics[width=1.1in,height=1.5in,clip,keepaspectratio]{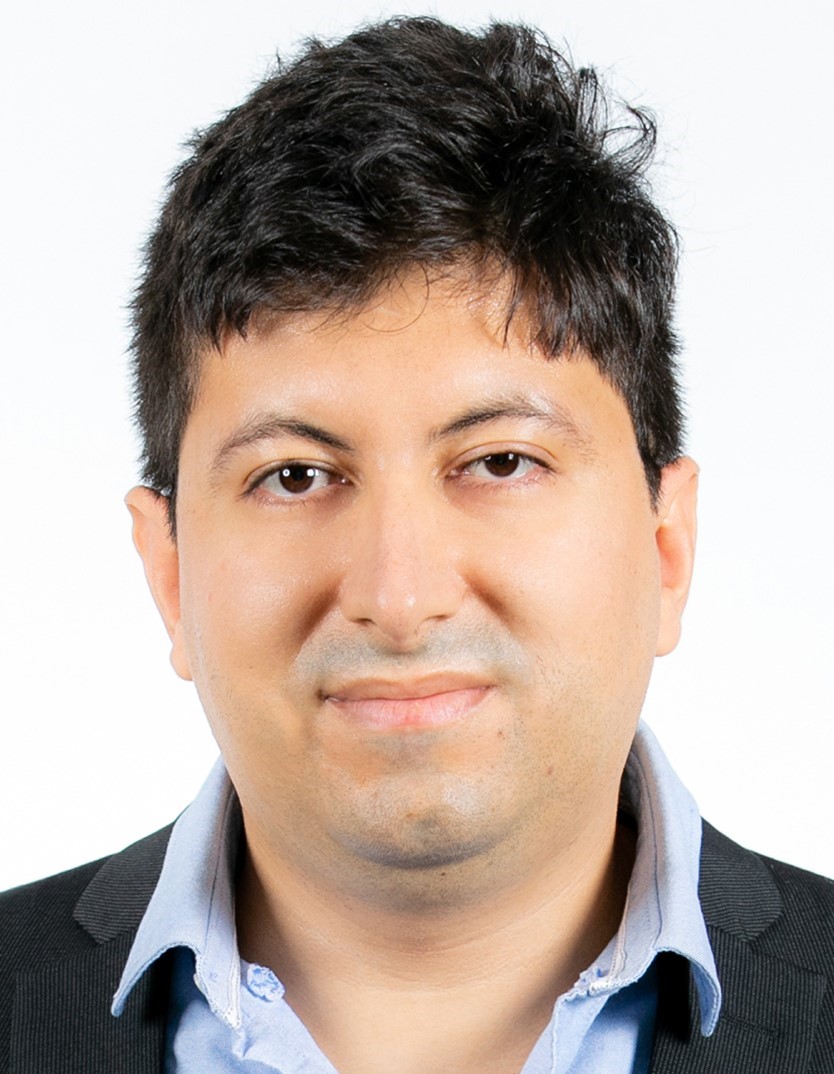}}]{Zine el abidine Kherroubi} received a B.S. degree in automatic control systems engineering from École Nationale Polytechnique d’Alger, Algiers, Algeria in 2015 and an M.S. degree in mobility and electric vehicles from Art et Métier ParisTech, Lille, France in 2016, as a Fellow Student of the prestigious Renault Foundation Scholarship. He received a Ph.D. degree in computer sciences from Claude Bernard Lyon 1 University, France in 2020, in joint research between the LIRIS laboratory and Renault car manufacturer. He worked as an R\&D engineer at Renault car manufacturer, France, between 2017 and 2020. He is currently with the Technology Innovation Institute, Abu Dhabi, UAE, as a researcher. His research interests include the application of AI/ML techniques for connected and autonomous vehicles technology, V2X network, collaborative driving, and vehicle-infrastructure cooperation.
\end{IEEEbiography}

\vspace{11pt}


\vfill

\end{document}